\documentclass[runningheads]{llncs}
\usepackage{graphicx}
\usepackage{amsmath,amssymb} 
\usepackage{color}

\usepackage{times}
\usepackage{epsfig}
\usepackage{graphicx}
\usepackage{amsmath}
\usepackage{amssymb}
\usepackage{bm}
\usepackage{xcolor}
\usepackage{tabu}
\usepackage{color,array}
\usepackage{multirow,makecell}
\usepackage{subfig}

\usepackage{cite}
\usepackage{blindtext,bm}
\usepackage{graphicx}

\usepackage[utf8]{inputenc}
\usepackage[english]{babel}

\usepackage{algorithm}
\usepackage{pdfpages}
\usepackage{colortbl}
\usepackage{array,multirow}
\usepackage{empheq}
\usepackage{relsize}
\usepackage{algorithmic}

\newcommand{\etal}{\textit{et al}. }
\newcommand{\ie}{\textit{i}.\textit{e}., }

\begin{document}
\title{Cross-View Kernel Similarity Metric Learning Using Pairwise Constraints for Person Re-identification} 

\titlerunning{Cross-View Kernel Similarity Metric Learning Using Pairwise Constraints}

\author{T M Feroz Ali \inst{1} \and Subhasis Chaudhuri\inst{1} }
\authorrunning{T M Feroz Ali and S. Chaudhuri}

\institute{Indian Institute of Technology Bombay, Mumbai, India \and
\email{\{ferozalitm,sc\}@ee.iitb.ac.in}}
\maketitle             
\begin{abstract}
Person re-identification is the task of matching pedestrian images across non-overlapping cameras. In this paper, we propose a non-linear cross-view similarity metric learning for handling small size training data in practical re-ID systems. The method  employs non-linear mappings combined with cross-view discriminative subspace learning and cross-view distance metric learning based on pairwise similarity constraints. It is a natural extension of XQDA from linear to non-linear mappings using kernels, and  learns non-linear transformations for efficiently handling complex non-linearity of person appearance across camera views. 
Importantly, the proposed method is very computationally efficient. Extensive experiments on four challenging datasets shows that our method attains competitive performance against state-of-the-art methods.

\keywords{ Person re-identification, Metric learning, Kernel XQDA}
\end{abstract}

\section{Introduction}

Person re-identification (re-ID) is the problem of matching person images from one camera view against the images captured from other non-overlapping camera views. Re-ID is a very challenging task as images of same person have significant appearance changes across views, due to large variation in illumination, background and pose. Also the low resolution surveillance cameras and common pedestrian attributes cause high visual similarity among different persons. 

Most existing methods for person re-identification concentrate on (i) design of identity discriminative feature descriptors  and (ii) distance metric learning. The hand crafted feature descriptors \cite{LOMO,GOG,LisantiPAMI14} have improved the re-ID performance, but they are alone insufficient in handling the large appearance changes across cameras. Hence the distance metric learning methods\cite{NK3ML,LOMO,rPcca,Zheng:nfst,IRS,SSSVM,MFML,SemiNK3ML} are used to learn a better similarity measure such that, irrespective of the view, same class samples are closer and distinct class samples are well separated. 

In recent years, though deep learning methods \cite{ImprDeep,MuDeep,PTGAN,Beyond:triplet_loss, DGD, TCP, SpindleNet, SLSTM, SCNN} have made good improvement in re-ID performance, they have a fundamental limitation in practical deployment as they need a large, annotated training data. 
Even with pre-trained networks, based on auxiliary/external supervision, such methods struggle to perform on small size training data.
Hence we refrain from using deep learning methods in this paper and instead concentrate on the following problem: "Given a \textit{small size training data} with given feature descriptors, can we design a better re-ID system, \textit{without} using any auxiliary/external supervision". 



Metric learning methods have shown a good performance in handling small size training data.  
 However, most of them have two fundamental limitations: 
\noindent (\textbf{I}) \textit{Small Sample Size (SSS) problem}: The SSS problem occurs when the number of training samples is less than the feature dimension. This creates singularity of inter/intra class scatter matrices. Hence most methods use unsupervised dimensionality reduction, which tend to make them sub-optimal. 
\noindent (\textbf{II}) \textit{Less Efficient Models}: Person appearance undergoes complex non-linear transformation across views.  However, most existing methods use an inherent linear transformation of the input features, which limits their capability in learning non-linear features. 

For addressing the above two limitations, we propose a new non-linear metric learning method, referred to as, \textit{Kernel Cross-view Quadratic Discriminant Analysis (k-XQDA)}.
It is a kernalized (non-linear) counterpart of XQDA\cite{LOMO}, which is one of the most popularly applied metric learning method in re-ID literature. k-XQDA uses mapping of the data samples to a very high dimensional kernel space, where it learns a cross-view distance metric and a cross-view discriminative subspace simultaneously, using pairwise similarity constraints. It is capable of learning highly effective non-linear features in the input feature space. k-XQDA efficiently handles the non-linearity in cross-view appearance and perform competitively against state-of-the-art methods. Importantly, our kernelized approach is computationally more efficient compared to the baseline methods.



\section{Related Methods}

%
%
%
%
%
%
%
%
%

Using given standard feature descriptors, the supervised metric learning methods generally learn a discriminative subspace or a Mahalanobis distance metric where the inter-class samples come closer and intra-class samples get well separated. 
The subspace learning methods like LFDA \cite{LFDA:CVPR}, NFST\cite{Zheng:nfst}, NK3ML \cite{NK3ML} and IRS \cite{IRS} use classification based model to learn discriminative features that generalize well to unseen data. 
For example, LFDA \cite{LFDA:CVPR} learned a discriminative subspace that maximize the ratio of between class variance and the within class variance, while preserving the local neighborhood structure of the data. NFST \cite{Zheng:nfst} used a more optimal discriminative nullspace to maximally collapse the same class samples to a single point. NK3ML \cite{NK3ML} and IRS \cite{IRS} were proposed to overcome the limitation of NFST in discriminating inter-class samples. The Mahalanobis distance metric based methods like LMNN \cite{LMNN1}, LDML \cite{LDML}, KISSME \cite{KISSME}, MLAPG \cite{MLAPG} 
 learn a Mahalanobis distance function of form $d(\mathbf{x}_i,\mathbf{x}_j)=(\mathbf{x}_i-\mathbf{x}_j)^T M(\mathbf{x}_i,-\mathbf{x}_j)$, where $M\succcurlyeq 0$ is a positive semi-definite matrix.  LDML \cite{LDML} used a probabilistic view for learning the Mahalanobis metric. LMNN \cite{LMNN1} learned the metric using constraints that ensure a margin between similar and dissimilar class samples. KISSME \cite{KISSME} considered the space of pairwise differences to define similar and dissimilar class, and then used a log likelihood ratio test to obtain a Mahalanobis distance metric. In order to take advantage of both subspace learning and Mahalanobis distance metric learning methods, S. Liao \etal proposed XQDA that simultaneously learned a cross-view discriminative subspace along with KISSME based cross-view distance metric. 
 
However, due to the large non-linearity in person appearance across cameras, the linear transformation induced by the above methods are unlikely to discriminate the persons efficiently. Hence kernel based distance metric learning methods \cite{rPcca,Zheng:nfst,IRS,kKISSME} were introduced to handle non-linearity in re-ID. F. Xiong \etal kernalized LFDA\cite{LFDA:CVPR} to obtain kLFDA\cite{rPcca}. Similarly L. Zhang \etal used kernel-NFST\cite{Zheng:nfst} and H. Wang \etal used the kernel-IRS\cite{IRS}. Recently kernalized version of KISSME, namely k-KISSME\cite{kKISSME} was derived and used to successfully improve the re-ID performance.


XQDA\cite{LOMO} is one the most popular metric learning methods in re-ID literature and has been used  in conjunction with many methods like GOG\cite{GOG}, SSDAL\cite{SSDAL}, SSM\cite{song:scalableManifold}, and also applied with recent deep learning based methods \cite{Reranking:kreciprocal}. However, it uses inherent linear transformation for learning the features. Hence obtaining an efficient kernalized (non-linear) version of XQDA becomes highly relevant. However, deriving the kernalized version of a method is not always a trivial task and may need complex analysis. In this paper, we derive the kernalized version of XQDA, namely k-XQDA. We show that k-XQDA can learn highly efficient non-linear features to handle the complex variations in person appearance. k-XQDA naturally handles SSS problem, since k-XQDA is a kernel based method, where the inherent matrices used in its computations have dimensions that are independent of feature dimensions and depends only on the training sample size. Our k-XQDA can handle small size training data effectively. We also show through our rigorous derivations, though involved, we finally attain simplified expressions that are computationally very efficient and fast, making it suitable for  practical implementation. 


\section{Kernel Cross-View Quadratic Discriminant Analysis}
We first revisit KISSME and XQDA. Then we present the proposed method k-XQDA.

\subsection{KISSME revisit}
KISSME learns distance metric based on equivalence constraints given as similar or dissimilar pairs. 
Given data samples $\mathbf{x} \in \mathbb{R}^d$ in the input feature space, belonging to $c$ classes, they consider the space of all pairwise sample differences $\Delta_{ij} = \mathbf{x}_i-\mathbf{x}_j$ and defines two classes, similar class $\Omega_S$ and 
dissimilar class $\Omega_D$,  containing $n_S$ and $n_D$ samples, respectively.  The pairwise difference would be comparatively small for similar class $\Omega_S$ samples and large for dissimilar class $\Omega_D$ samples.  By distinguishing the variations of the two classes, any general multiclass classification problem is subsequently solved. 
As the pairwise differences are symmetric, both the classes  $\Omega_S$ and  $\Omega_D$ are assumed to be zero mean Gaussian distributions with covariance $\Sigma_S$ and $\Sigma_D$. Motivated by statistical inference perspective, the optimal decision function $\delta(\Delta_{ij})$ that indicates whether a difference pair $\Delta_{ij}$ belongs to the similar or dissimilar class is obtained by a log likelihood ratio test of the two Gaussian distributions. 
\begin{eqnarray}
\delta(\Delta_{ij}) &=&log \Big(\frac{p(\Delta_{ij}|\Omega_D)}{p(\Delta_{ij}|\Omega_S)}\Big)\\
&=&log\Bigg(\frac{\frac{1}{(2\pi)^{d/2}|\Sigma_D|}exp(-\frac{1}{2} \Delta_{ij}^T \Sigma_D^{-1}\Delta_{ij})}{\frac{1}{(2\pi)^{d/2}|\Sigma_S|}exp(-\frac{1}{2} \Delta_{ij}^T \Sigma_S^{-1}\Delta_{ij})}\Bigg)
\end{eqnarray}
A high value of $\Delta_{ij}$ implies that $\Delta_{ij} \in \Omega_D$, while a low value implies $\Delta_{ij} \in \Omega_S$.
The decision function is simplified \cite{KISSME} to get
\begin{eqnarray}
\delta(\Delta_{ij}) &\propto& \Delta_{ij}^T(\Sigma_S^{-1}-\Sigma_D^{-1})\Delta_{ij} \, ,
\end{eqnarray}
and finally the KISSME distance metric is obtained that mirror the properties of the log likelihood ratio test, as given below. 
\begin{eqnarray}
d(\mathbf{x}_i,\mathbf{x}_j) &=& (\mathbf{x}_i-\mathbf{x}_j)^T (\Sigma_S^{-1}-\Sigma_D^{-1})_{+} (\mathbf{x}_i-\mathbf{x}_j)
\label{eqn:KISSME}
\end{eqnarray}
where $(\cdot)_{+}$ represents the projection to the cone of positive semi-definite matrices using eigen analysis, to ensure  (\ref{eqn:KISSME}) to be a valid Mahalanobis distance metric. It can be seen that learning the KISSME distance metric corresponds to estimating the covariance matrices $\Sigma_S$ and $\Sigma_D$.
\begin{eqnarray}
\Sigma_S =\sum_{\Delta_{ij} \in \Omega_S} (\mathbf{x}_{i}-\mathbf{x}_j)(\mathbf{x}_{i}-\mathbf{x}_j)^T \nonumber \\
 \Sigma_D =\sum_{\Delta_{ij} \in \Omega_D} (\mathbf{x}_{i}-\mathbf{x}_j)(\mathbf{x}_{i}-\mathbf{x}_j)^T
\label{KISSME_CovMat_calc}
\end{eqnarray}

\subsection{XQDA revisit}
KISSME becomes intractable in very high dimensions and hence it uses PCA on the input features to get a low dimensional subspace, where $\Sigma_S$ and $\Sigma_D$ are estimated. However, the unsupervised dimensionality reduction doesn't consider distance metric learning and can loose discriminative information. Also KISSME considers single view data, \ie  
it does not account any distinction of camera views for considering the pairwise sample differences.

In order to address the above two limitations, S. Liao \etal extended KISSME and proposed a \textit{cross-view} metric learning approach called Cross-view Quadratic Discriminant Analysis (XQDA), where cross view data is used to learn a cross view discriminative subspace and a cross-view similarity measure simultaneously.  


In particular, given samples from $c$ classes, with $n$ samples $\mathbf{X} = (\mathbf{x}_1, \mathbf{x}_2, \ldots,\mathbf{x}_n)$ from one view and $m$ samples $\mathbf{Z} = (\mathbf{z}_1, \mathbf{z}_2, \ldots,\mathbf{z}_m)$ from the other view, s.t. $\mathbf{x}_i, \mathbf{z}_i \in \mathbb{R}^d$, XQDA uses cross-view training set $\{\mathbf{X},\mathbf{Z}\}$ and considers  the $nm$ pairwise sample differences \textit{across views} to estimate the cross-view similar and dissimilar classes, making the distance metric more viewpoint invariant.  XQDA learns a subspace $W = (\mathbf{w}_1,\mathbf{w}_2,\ldots, \mathbf{w}_b) \in \mathbb{R}^{d \times b}$ that maximize the discrimination between the two classes $\Omega_S$ and 
 $\Omega_D$, and learn a distance measure, similar to Eq. (\ref{eqn:KISSME}), as 
\begin{equation}
d(\mathbf{x}_{i},\mathbf{z}_{j}) = (\mathbf{x}_{i}-\mathbf{z}_{j})^T W(\Sigma^{\prime-1}_{S} - \Sigma^{\prime-1}_{D})_{+} W^T(\mathbf{x}_{i}-\mathbf{z}_{j})
\label{eqn:XQDAmetric}
\end{equation}  
where $\Sigma^{\prime}_{S} = W^T \Sigma_{S}W$, $\Sigma^{\prime}_{D} = W^T \Sigma_{D}W$. 
As the classes $\Omega_S$ and $\Omega_D$ have zero mean, Fisher criterion based LDA can not be directly used to learn the subspace $W$ that discriminates the classes. However, XQDA uses the class variances $\sigma_S$ and $\sigma_D$ to discriminate the classes. More specifically, XQDA obtains the discriminant vectors $\mathbf{w}_k$ in $W$ such that they maximize the ratio of the class variances $\sigma_D(\mathbf{w}_k)$ and $\sigma_S(\mathbf{w}_k)$, in the corresponding directions, which has a form of Generalized Rayleigh Quotient,
\begin{eqnarray}
J(\mathbf{w}_k) =  \frac{\sigma_D(\mathbf{w}_k)}{\sigma_S(\mathbf{w}_k)} = \frac{\mathbf{w}^T_k \Sigma_{D}\mathbf{w}_k}{\mathbf{w}^T_k \Sigma_{S}\mathbf{w}_k}\,.
\label{eqn:XQDAcost}
\end{eqnarray}
Thus XQDA finds the subspace $W$ such that the variance of $\Omega_D$ is maximized, while variance of $\Omega_S$ is minimized, thereby discriminating the two class based on their variances. The optimal discriminants are composed of the eigenvectors corresponding to  $b$ largest eigenvalues of $\Sigma_{S}^{-1}\Sigma_{D}$.\\

\textbf{Efficient Computation}:
As there are $nm$ pairwise sample differences, the calculation of cross-view covariance matrices $\Sigma_{D}$ and $\Sigma_{S}$ using (\ref{KISSME_CovMat_calc}) requires $\mathcal{O}(mnd^2)$ and $\mathcal{O}(NKd^2)$, multiplications respectively, where $N=max(m,n)$ and $K$ is the average number of samples per class. However, the covariance matrices can be efficiently calculated without actually computing the $nm$ pairwise differences, by simplifying them as follows:
 \begin{align}
n_S \Sigma_S &= \widetilde{\mathbf{X}}\widetilde{\mathbf{X}}^T + \widetilde{\mathbf{Z}}\widetilde{\mathbf{Z}}^T - \mathbf{S}\mathbf{R}^T -\mathbf{R}\mathbf{S}^T \label{eqn:SigmaS}\\		
n_D \Sigma_D &= m \mathbf{X}\mathbf{X}^T + n \mathbf{Z}\mathbf{Z}^T -\mathbf{s}\mathbf{r}^T -\mathbf{r}\mathbf{s}^T -n_S\Sigma_S 
\label{eqn:SigmaD}
 \end{align} 	  
where $\widetilde{\mathbf{X}} = (\sqrt{m_1}\mathbf{x}_1, \sqrt{m_1}\mathbf{x}_2, \ldots, \sqrt{m_1}\mathbf{x}_{n_1}, \ldots, \sqrt{m_c}\mathbf{x}_n)$, 
$\widetilde{\mathbf{Z}} = (\sqrt{n_1}\mathbf{z}_1, \sqrt{n_1}\mathbf{z}_2, \ldots,$ $\sqrt{n_1}\mathbf{z}_{m_1} \ldots, \sqrt{m_c}\mathbf{z}_m),
\;\;\;  \mathbf{S} = (\sum_{{y_i}=1}\mathbf{x}_i, \sum_{{y_i}=2}\mathbf{x}_i, \ldots, \sum_{{y_i}=c}\mathbf{x}_i), \;\;\;\;  {\mathbf{s}=\sum_{i=1}^n \mathbf{x}_i,} \\
{\mathbf{R} = (\sum_{{y_j}=1}\mathbf{z}_j, \sum_{{y_j}=2}\mathbf{z}_j, \ldots, \sum_{{y_j}=c}\mathbf{z}_j)}, \;\;\;{\mathbf{r} =  \sum_{j=1}^m \mathbf{z}_j}$, $y_i,y_j \in \{1,\ldots,c\}$ are the class labels of $\mathbf{x}_i$ and $\mathbf{z}_j$ respectively, $n_i$ is the number of samples for class $y_i$ in $\mathbf{X}$ and $m_i$ is the number of samples for class $y_j$ from $\mathbf{Z}$. The simplified expressions in (\ref{eqn:SigmaS}) and (\ref{eqn:SigmaD}), reduces the computations of both the covariance matrices to $\mathcal{O}(Nd^2)$.

\subsection{Kernel-XQDA}
Next, we propose how XQDA can be kernalized to obtain its non-linear version \mbox{k-XQDA}. Kernel methods use a non-linear mapping of input samples to a high dimensional space, implicitly determined by a kernel function. In the kernel space, the primary model and the inherent transformations are learned, which results in learning the corresponding non-linear models and transformations in the input feature space. 

Let the  kernel function be $k(\mathbf{x}_i,\mathbf{x}_j)= \langle\phi(\mathbf{x}_i),\phi(\mathbf{x}_j)\rangle$, where $\phi(\mathbf{x})$ is the non-linear mapping of the input sample $\mathbf{x}$ to the high dimensional kernel space $\mathcal{F}$. 
 For kernalization, the XQDA model has to be formulated in terms of inner products $ \langle\phi(\mathbf{x}_i),\phi(\mathbf{x}_j)\rangle$, which is then replaced using the kernel function $k(\mathbf{x}_i,\mathbf{x}_j)$. Hence the derivation of k-XQDA involves mainly (\textbf{I}) the kernalization of the cost function $J(\mathbf{w}_k)$ in (\ref{eqn:XQDAcost}) and (\textbf{II}) the distance metric function $d(\mathbf{x}_i,\mathbf{z}_j)$ in (\ref{eqn:XQDAmetric}).

Note that the kernelization of the cost function (\ref{eqn:XQDAcost}) involves kernelizing w.r.t the covariance matrices, for which, a clean and straightforward way is to use the expressions in (\ref{KISSME_CovMat_calc}), based on indexing.
However, it would require computing the outer product for $nm$ pairwise differences, making k-XQDA computationally inefficient. Hence we strictly adhere to use the expressions in (\ref{eqn:SigmaS}) and (\ref{eqn:SigmaD}) itself, in order to make k-XQDA computationally efficient.  However, kernelizing using the later is a complex task mainly due to two reasons: (i) The matrices $\widetilde{\mathbf{X}}, \mathbf{S}, \mathbf{X}, \mathbf{s}$ depends on data samples from one view, while the matrices   $\widetilde{\mathbf{Z}}, \mathbf{R}, \mathbf{Z}, \mathbf{r}$ depends on the data samples from the other view. Hence we need to separately account the kernel functions corresponding to each view. (ii)Computing the kernel functions corresponding to $\mathbf{S}, \mathbf{R}, \mathbf{s}, \mathbf{r}$ involves separately computing the kernel functions for the mean of each class and all classes from each view. However, we show that, though the derivations are little involved, we finally obtain clean and elegant kernelized expressions for the covariance matrices and the cost function (\ref{eqn:XQDAcost}), which are also computationally very efficient for practical implementation.

 Given the cross-view training data $(\mathbf{X},\mathbf{Z})\in \mathbb{R}^{d \times (n+m)}$, the kernel matrix $\mathbf{K} \in \mathbb{R}^{(n+m) \times (n+m)}$ can be calculated and expressed as block matrices of the form
\begin{eqnarray}
\mathbf{K}=\left[\begin{array}{@{}c|c@{}}
   K_{XX} & K_{XZ}\\
   \hline
   K_{ZX} & K_{ZZ}
    \end{array}\right]
    \label{eqn:mainK}
\end{eqnarray}
where the block-matrices $K_{XX} \in \mathbb{R}^{n \times n}$, $K_{ZZ} \in \mathbb{R}^{m \times m}$, $K_{XZ} \in \mathbb{R}^{n \times m}$ and $K_{ZX} \in \mathbb{R}^{m \times n}$ are such that
\begin{equation}
K_{XX}=\Phi_X^T\Phi_X,\;K_{ZZ}=\Phi_Z^T\Phi_Z,\;K_{XZ}=\Phi_X^T\Phi_Z, \;K_{ZX}=\Phi_Z^T\Phi_Z
\label{eqn:Kblockmatppty}
\end{equation} 
Note that each of the block matrices $K_{XX}$ and $K_{ZZ}$ are the kernel matrices corresponding to the samples of separate views, and the block matrices $K_{XZ}$ and $K_{ZX}$ are the kernel matrices corresponding to the samples across views. Also the block matrices have the following symmetry properties:
\begin{equation}
K_{XX} = K_{XX}^T, \quad K_{ZZ} = K_{ZZ}^T, \quad  K_{XZ} = K_{ZX}^T.
\label{eqn:Symmppty}
\end{equation}
In the kernel space $\mathcal{F}$, every discriminant vector $\mathbf{w}_k$ lies in the span of the training data set $\{\phi(\mathbf{x}_1),\ldots, \phi(\mathbf{x}_n), \phi(\mathbf{z}_1), \ldots, \phi(\mathbf{z}_m)\}$. Hence $\mathbf{w}_k$ can be expressed in the form:
\begin{eqnarray}
\mathbf{w}_k &=& \sum_{i=1}^n \alpha_i^{(k)} \phi(\mathbf{x}_i) + \sum_{j=1}^m \beta_j^{(k)} \phi(\mathbf{z}_j) 
\label{eqn:rep_theorem}
\end{eqnarray}
It should be noted that in conventional kernel methods, a vector $\mathbf{w}$ in the feature space $\mathcal{F}$ is expressed using expansion coefficients $\alpha$ as $\mathbf{w} = \sum_i \alpha_i^{(k)} \phi(\mathbf{x}_i)$. However, in (\ref{eqn:rep_theorem}) we use two expansion coefficients $\alpha$ and $\beta$, in order to separately account the samples belonging to each view. The vector $\mathbf{w}_k$ in (\ref{eqn:rep_theorem}) can be rewritten as

\begin{eqnarray}
\mathbf{w}_k &=& \Phi_X \bm{\alpha}_k + \Phi_Z \bm{\beta}_k =   \bm{\Phi}\bm{\theta}_k
\label{eqn:rep_theorem2}
\end{eqnarray}
where $\Phi_{X} = [\phi(\mathbf{x}_1),\dots,\phi(\mathbf{x}_n)]$ and $\Phi_{Y} = [\phi(\mathbf{z}_1), \dots, \phi(\mathbf{z}_m)]$ are respectively the matrix functions that map all the samples of $\mathbf{X}$ and $\mathbf{Z}$ to the kernel space $\mathcal{F}$, and $\bm{\alpha}_k = [\alpha_1^{(k)}, \alpha_2^{(k)}, \ldots, \alpha_n^{(k)}]^T$ and $\bm{\beta}_k = [\beta_1^{(k)}, \beta_2^{(k)}, \ldots, \beta_m^{(k)}]^T$ are the expansion coefficient vectors corresponding to each view, $ \bm{\theta}_k = \left[\bm{\alpha}_k,\bm{\beta}_k\right]^T$ is the combined expansion coefficient vector and $\bm{\Phi} = [\Phi_{X}, \Phi_{Z}]$ . Hence  $ \mathbf{w}_k$ in the kernel space is represented using $\bm{\alpha}_k$ and $\bm{\beta}_k$, or equivalently by $\bm{\theta}_k$.

In the following we show how XQDA's cost function $J(\mathbf{w}_k)$ in (\ref{eqn:XQDAcost}) and the distance metric $d(\mathbf{x}_i,\mathbf{z}_j)$ in (\ref{eqn:XQDAmetric}) can be kernelized: \\

\noindent \textbf{3.3.1   $\quad$  Kernelization of cost function $J(\mathbf{w}_k)$:}\\

\noindent We show that both the numerator term $\mathbf{w}^T_k \Sigma_{D}\mathbf{w}_k$ and denominator term $\mathbf{w}^T_k \Sigma_{S}\mathbf{w}_k$ of the cost function $J(\mathbf{w}_k)$ can be formulated in terms of inner products and hence they can be separately kernalized.


\vspace{2mm}
\noindent \textbf{Kernelization of denominator} $\mathbf{w}^T_k \Sigma_{S}\mathbf{w}_k$: 
As seen in Eq.(\ref{eqn:SigmaS}), $\Sigma_{S}$ is a function of $\widetilde{\mathbf{X}},\widetilde{\mathbf{Z}},\mathbf{S},\mathbf{R}$, which are in turn functions of the training set samples. So we first express these matrices in the kernel space $\mathcal{F}$ using the function $\phi(\cdot)$ as follows:

\begin{eqnarray}
\Phi_{\widetilde{X}} &=& [\sqrt{m_1}\phi(\mathbf{x}_1),\dots,\sqrt{m_1}\phi(\mathbf{x}_{n_1}),\ldots,\sqrt{m_c}\phi(\mathbf{x}_n)] \label{eqn:PhiXtilde} \\
\Phi_{\widetilde{Z}} &=& [\sqrt{n_1}\phi(\mathbf{z}_1),  \dots, \sqrt{n_1}\phi(\mathbf{z}_{m_1}), \dots, \sqrt{n_c}\phi(\mathbf{z}_m)] \label{eqn:PhiZtilde}\\
\Phi_{S} &=& (\sum_{{y_i}=1}\phi(\mathbf{x}_i), \sum_{{y_i}=2}\phi(\mathbf{x}_i), \ldots, \sum_{{y_i}=c}\phi(\mathbf{x}_i)) \label{eqn:PhiS}\\
\Phi_{R} &=& (\sum_{{y_j}=1}\phi(\mathbf{z}_j), \sum_{{y_j}=2}\phi(\mathbf{z}_j), \ldots, \sum_{{y_j}=c}\phi(\mathbf{z}_j))\label{eqn:PhiR}
\end{eqnarray}
Then, using (\ref{eqn:SigmaS}), the covariance matrix $\Sigma_S$ in $\mathcal{F}$ can be expressed as
\begin{eqnarray}
n_S \Sigma_S &=& \underbrace{\Phi_{\widetilde{X}}\Phi_{\widetilde{X}}^T}_{A} 
+ \underbrace{\Phi_{\widetilde{Z}}\Phi_{\widetilde{Z}}^T}_{B} -\underbrace{\Phi_{S}\Phi_{R}^T}_{C} -\underbrace{\Phi_{R}\Phi_{S}^T}_{D}
\label{eqn:ABCD}
\end{eqnarray}  
Then using Eq. (\ref{eqn:rep_theorem2}) and (\ref{eqn:ABCD}), the numerator term $\mathbf{w}_k^T  \Sigma_S \mathbf{w}_k $ can be written as
\begin{align}
\mathbf{w}_k^Tn_S \Sigma_S \mathbf{w}_k =   
 f_A(\bm{\alpha}_k,\bm{\beta}_k) + f_B(\bm{\alpha}_k,\bm{\beta}_k)+f_C(\bm{\alpha}_k,\bm{\beta_k})+f_D(\bm{\alpha}_k,\bm{\beta}_k) \label{eqn:Num}
\end{align} 
where the functions $f_A$, $f_B$, $f_C$ and $f_D$ are of the form
\begin{align}  
f_Y(\bm{\alpha}_k,\bm{\beta}_k) &= \bm{\alpha}_k^T \Phi_{X}^T Y \Phi_{X} \bm{\alpha}_k +
\bm{\beta}_k^T \Phi_{Z}^T Y \Phi_{Z} \bm{\beta}_k  \nonumber \\
& \qquad  +\bm{\alpha}_k^T \Phi_{X}^T Y \Phi_{Z} \bm{\beta}_k +
\bm{\beta}_k^T \Phi_{Z}^T Y \Phi_{X} \bm{\alpha}_k \label{eqn:f_A}
\end{align}  
for $Y=A,B,C,D$, which are defined in (\ref{eqn:ABCD}). Next we show that each of the functions in (\ref{eqn:Num}) can be expressed in terms of inner products of $\Phi$ and hence can be individually kernelized. We have the following Lemmas.\\

\noindent \textit{ \textbf{Lemma 1:} 
$f_A(\bm{\alpha},\bm{\beta})$ can be kernalized as 
$f_A(\bm{\alpha}_k,\bm{\beta}_k) = \bm{\theta}_k^T\widetilde{A}\bm{\theta}_k$, where
\begin{eqnarray}
\widetilde{A} &=  	\left[\begin{array}{@{}cc@{}}
K_{XX}\\
K_{ZX}
  \end{array}\right]
  \left[\begin{array}{@{}c@{}}
\widetilde{F}
   \end{array}\right]   
    \left[\begin{array}{@{}cc@{}}
K_{XX} & K_{XZ}
   \end{array}\right] \, , 
   \label{eqn:Atilde}
  \end{eqnarray}
$\widetilde{F} = \text{diag}(m_1 I_{n_1},m_2 I_{n_2},\ldots,m_c I_{n_c}) \in \mathcal{R}^{n \times n}$, such that $I_{n_i}$ is identity matrix of size $(n_i \times n_i)$.}\\

\noindent\textit{\textbf{Proof:}} We have $A = \Phi_{\widetilde{X}}\Phi_{\widetilde{X}}^T $. However, for kernelization of A, we need to express it in terms of $\Phi_{X}$, which is not trivial due to the presence of coefficients $\sqrt{m_1},\ldots, \sqrt{m_c}$, as seen in (\ref{eqn:PhiXtilde} ). In order to decouple the coefficients, we do the following. Let $\widetilde{F}$ be a diagonal matrix defined as $\widetilde{F} = \text{diag}(m_1 I_{n_1},m_2 I_{n_2},\ldots,m_c I_{n_c})  \in \mathbb{R}^{n \times n}$, \ie
	\begin{eqnarray}\small
	\widetilde{F} = 
  	\left(\begin{array}{@{}ccc|ccc|c|ccc@{}}
    m_1 &  &  &  &  & &&&&\\
    & \ddots &  &  &  & &&&&\\
    && m_1 &  &  &  &&&&\\
    \cline{1-6}
         &&& m_2 &  &  &  &  &&\\
    &&& & \ddots &  &  &  &&\\
    &&& && m_2 &  &  &&\\
     \cline{4-7}
      &&& &&& \ddots &  & & \\
           \cline{7-10}
    &&& &&& &m_c &   & \\
    &&& &&& && \ddots  & \\
    &&& &&& && &m_c \\
  \end{array}\right)
\label{eqn:Ftilde}
\end{eqnarray}  
where, $m_j$ is the number of samples for class $y_j$ from $\mathbf{Z}$.
Then, using (\ref{eqn:PhiXtilde}) and the definition of the matrix $A$, it can be factorized in terms of $\Phi_X$ using the decoupling matrix $\widetilde{F}$ as follows:
	\begin{eqnarray}
	A = \Phi_{\widetilde{X}}\Phi_{\widetilde{X}}^T 
	=\Phi_X  \widetilde{F}  \Phi_X^T \label{eqn:A}
	\end{eqnarray}	
Then using Eq. (\ref{eqn:f_A}), (\ref{eqn:A}) and (\ref{eqn:Kblockmatppty}), we can express $f_A(\bm{\alpha},\bm{\beta})$ in terms of inner products of $\Phi$ and later kernelize as shown below:

\begin{align*}
	f_A(\bm{\alpha}_k,\bm{\beta}_k)		
&= \bm{\alpha}_k^T \Phi_X^T A \Phi_X \bm{\alpha}_k +
\bm{\beta}_k^T \Phi_Z^T A \Phi_Z \bm{\beta}_k \nonumber+ \bm{\alpha}_k^T \Phi_X^T A \Phi_Z \bm{\beta}_k +
\bm{\beta}_k^T \Phi_Z^T A \Phi_X \bm{\alpha}_k \\
	&= \bm{\alpha}_k^T \Phi_X^T \Phi_X 	\widetilde{F} \Phi_X^T \Phi_X  \bm{\alpha}_k + 
	 \bm{\beta}_k^T \Phi_Z^T \Phi_X \widetilde{F} 	\Phi_X^T \Phi_Z  \bm{\beta}_k  \\ & \qquad + \bm{\alpha}_k^T \Phi_X^T \Phi_X \widetilde{F} 				\Phi_X^T \Phi_Z  \bm{\beta}_k 
	 + \bm{\beta}_k^T \Phi_Z^T \Phi_X \widetilde{F} \Phi_X^T \Phi_X  \bm{\alpha}_k \\	
	 	&= \bm{\alpha}_k^T K_{XX} \widetilde{F} K_{XX} \bm{\alpha}_k +
	\bm{\beta}_k^T K_{ZX} \widetilde{F} K_{XZ}  \bm{\beta}_k \\ 
	& \qquad + \bm{\alpha}_k^T K_{XX} \widetilde{F} K_{XZ} \bm{\beta}_k  
	 + \bm{\beta}_k^T K_{ZX} \widetilde{F} K_{XX} \bm{\alpha}_k\\
	&= [\bm{\alpha}_k^T  \bm{\beta}_k^T ]
	\left[\begin{array}{@{}cc@{}}
K_{XX}\widetilde{F}K_{XX} & K_{XX}\widetilde{F}K_{XZ}\\
K_{ZX}\widetilde{F}K_{XX} & K_{ZX}\widetilde{F}K_{XZ}
  \end{array}\right]
  \left[\begin{array}{@{}c@{}}
  \bm{\alpha}_k \\
  \bm{\beta}_k
   \end{array}\right]\\
   	&= [\bm{\alpha}_k^T  \bm{\beta}_k^T ]
\left[\begin{array}{@{}cc@{}}
K_{XX}\\
K_{ZX}
  \end{array}\right]
  \left[\begin{array}{@{}c@{}}
\widetilde{F}
   \end{array}\right]   
    \left[\begin{array}{@{}cc@{}}
K_{XX} & K_{XZ}
   \end{array}\right] 
  \left[\begin{array}{@{}c@{}}
  \bm{\alpha}_k \\
  \bm{\beta}_k
   \end{array}\right]\\
   &=\bm{\theta}_k^T\widetilde{A}\bm{\theta}_k
\end{align*} {\hfill$\square$}


\noindent \textit{\textbf{Lemma 2:} 
$f_B(\bm{\alpha}_k,\bm{\beta}_k)$ can be kernalized  as 
$f_B(\bm{\alpha}_k,\bm{\beta}_k) = \bm{\theta}_k^T\widetilde{B}\bm{\theta}_k$, where
\begin{eqnarray}
\widetilde{B} 
&=\left[\begin{array}{@{}cc@{}}
K_{XZ}\\
K_{ZZ}
  \end{array}\right]
  \left[\begin{array}{@{}c@{}}
\widetilde{G}
   \end{array}\right]   
    \left[\begin{array}{@{}cc@{}}
K_{ZX} & K_{ZZ}
   \end{array}\right] 
      \label{eqn:Btilde}
\end{eqnarray}  
and $\widetilde{G} = \text{diag}(n_1 I_{m_1},n_2 I_{m_2},\ldots,n_c I_{m_c}) \in \mathcal{R}^{m \times m}$, such that $I_{m_i}$ is identity matrix of size $(m_i \times m_i)$.\\
}\\
\noindent \textit{\textbf{Proof:}} The kernelization of $f_B(\bm{\alpha}_k,\bm{\beta}_k)$ is similar to that of $f_A(\bm{\alpha}_k,\bm{\beta}_k)$. As $\widetilde{B} = \Phi_{\widetilde{Z}}\Phi_{\widetilde{Z}}^T$, we need to express it in terms of $\Phi_{Z}$ for kernelization, which is not directly possible as $\Phi_{\widetilde{Z}}$ is coupled with the coefficients $\sqrt{n_1},\ldots, \sqrt{n_c}$ (refer (\ref{eqn:PhiZtilde})). Hence we use a decoupling matrix $\widetilde{G}$ as follows. Let $\widetilde{G}$ be a diagonal matrix defined as $\widetilde{G} = \text{diag}(n_1 I_{m_1},n_2 I_{m_2},\ldots,n_c I_{m_c})  \in \mathbb{R}^{m \times m}$,\ie,
	\begin{eqnarray}\small
	\widetilde{G} = 
  	\left(\begin{array}{@{}ccc|ccc|c|ccc@{}}
    n_1 &  &  &  &  & &&&&\\
    & \ddots &  &  &  & &&&&\\
    && n_1 &  &  &  &&&&\\
    \cline{1-6}
         &&& n_2 &  &  &  &  &&\\
    &&& & \ddots &  &  &  &&\\
    &&& && n_2 &  &  &&\\
     \cline{4-7}
      &&& &&& \ddots &  & & \\
           \cline{7-10}
    &&& &&& &n_c &   & \\
    &&& &&& && \ddots  & \\
    &&& &&& && &n_c \\
  \end{array}\right)
  \label{eqn:Gtilde}
\end{eqnarray} 
where, $n_i$ is the number of samples for class $y_i$ from $\mathbf{X}$.
Then, using (\ref{eqn:PhiZtilde}), the decoupling matrix $\widetilde{G}$ and the definition of $B$,  the later can be factorized in terms of $\Phi_Z$ as follows:
	\begin{eqnarray}
	B = \Phi_{\widetilde{Z}}\Phi_{\widetilde{Z}}^T 
	=\Phi_Z  \widetilde{G}  \Phi_Z^T \label{eqn:B}
	\end{eqnarray}
Then using  (\ref{eqn:f_A}), (\ref{eqn:B}) and (\ref{eqn:Kblockmatppty}), we can kernelize $f_B(\bm{\alpha},\bm{\beta})$ as shown below:
\begin{align*}
f_B(\bm{\alpha}_k,\bm{\beta}_k)  &= \bm{\alpha}_k^T \Phi_X^T B \Phi_X 			\bm{\alpha}_k + 
	\bm{\beta}_k^T \Phi_Z^T B \Phi_Z 						\bm{\beta}_k 
	+	\bm{\alpha}_k^T \Phi_X^T B \Phi_Z 					\bm{\beta}_k  	 
	 + \bm{\beta}_k^T \Phi_Z^T B \Phi_X 						\bm{\alpha}_k\\	
 &=  \bm{\alpha}_k^T \Phi_X^T \Phi_Z 		\widetilde{G} \Phi_Z \Phi_X^T  \bm{\alpha}_k +
	\bm{\beta}_k^T \Phi_Z^T \Phi_Z \widetilde{G} 				\Phi_Z \Phi_Z^T  \bm{\beta}_k \\
	& \qquad + \bm{\alpha}_k^T \Phi_X^T \Phi_Z \widetilde{G} 				\Phi_Z \Phi_Z^T  \bm{\beta}_k 
	 + \bm{\beta}_k^T \Phi_Z^T \Phi_Z \widetilde{G} 				\Phi_Z \Phi_X^T  \bm{\alpha}_k \\	
&=  \bm{\alpha}_k^T K_{XZ} \widetilde{G} K_{ZX}^T  \bm{\alpha}_k + 
	\bm{\beta}_k^T K_{ZZ} \widetilde{G} K_{ZZ}^T  \bm{\beta}_k \\
	& \qquad + \bm{\alpha}_k^T K_{XZ} \widetilde{G} K_{ZZ}^T  \bm{\beta}_k 
  +  \bm{\beta}_k^T K_{ZZ} \widetilde{G} K_{ZX}^T  \bm{\alpha}_k\\
   &=  [\bm{\alpha}_k^T  \bm{\beta}_k^T ]
	\left[\begin{array}{@{}cc@{}}
K_{XZ}\widetilde{G}K_{ZX}^T & K_{XZ}\widetilde{G}K_{ZZ}^T\\
K_{ZZ}\widetilde{G}K_{XZ}^T & K_{ZZ}\widetilde{G}K_{ZZ}^T
  \end{array}\right]
  \left[\begin{array}{@{}c@{}}
  \bm{\alpha}_k \\
  \bm{\beta}_k
   \end{array}\right]\\
&=  [\bm{\alpha}_k^T  \bm{\beta}_k^T ]
	\left[\begin{array}{@{}cc@{}}
K_{XZ}\\
K_{ZZ}
  \end{array}\right]
  \left[\begin{array}{@{}c@{}}
\widetilde{G}
   \end{array}\right]   
    \left[\begin{array}{@{}cc@{}}
K_{ZX} & K_{ZZ}
   \end{array}\right] 
  \left[\begin{array}{@{}c@{}}
  \bm{\alpha}_k \\
  \bm{\beta}_k
   \end{array}\right]\\
   &= \bm{\theta}_k^T\widetilde{B}\bm{\theta}_k.
\end{align*}{\hfill$\square$}


Next, in order to kernelize $f_C(\bm{\alpha}_k,\bm{\beta}_k) $ and $f_D(\bm{\alpha}_k,\bm{\beta}_k)$, we define the following matrices. 
\begin{equation}
H_{XX} = \Phi_{X}^T \Phi_{S},\quad H_{ZZ} = \Phi_{Z}^T \Phi_{R}, \quad H_{XZ} = \Phi_{X}^T \Phi_{R}, \quad H_{ZX} = \Phi_{Z}^T \Phi_{S}
\label{eqn:H1}
\end{equation}
The above matrices are of size $H_{XX},H_{XZ} \in \mathcal{R}^{n \times c}$ and  $H_{ZX},H_{ZZ} \in \mathcal{R}^{m \times c}$. The $(p,q)$th element of each of these matrices can be expressed in terms of the kernel function $k(\mathbf{x}_i,\mathbf{x}_j)$ as 
\begin{eqnarray}
(H_{XX})_{pq} = \sum_{y_i=q} k(x_p,x_i), \;(H_{ZZ})_{pq} = \sum_{y_j=q} k(z_p,z_j) \nonumber\\
(H_{XZ})_{pq} = \sum_{y_j=q} k(x_p,z_j), \;(H_{ZX})_{pq} = \sum_{y_i=q} k(z_p,x_i)
\label{eqn:H2}
\end{eqnarray}
Then, we have the below Lemma.\\

\noindent\textit{\textbf{Lemma 3:} 
$f_C(\bm{\alpha},\bm{\beta})$ and $f_D(\bm{\alpha}_k,\bm{\beta}_k)$ can be kernalized  such that 
$f_C(\bm{\alpha}_k,\bm{\beta}_k) = \bm{\theta}^T_k\widetilde{C}\bm{\theta}_k$ and $f_D(\bm{\alpha},\bm{\beta}) = \bm{\theta}^T_k\widetilde{C}^T\bm{\theta}_k$, where}
\begin{eqnarray}
	\widetilde{C} = \left[\begin{array}{@{}cc@{}}
H_{XX}\\
H_{ZX}
  \end{array}\right]
    \left[\begin{array}{@{}cc@{}}
H_{XZ}^T & H_{ZZ}^T
   \end{array}\right] 
   \label{eqn:C}
\end{eqnarray}  

\noindent\textit{\textbf{Proof:} }
Using (\ref{eqn:f_A}), the relations in (\ref{eqn:H1}) and the definition $C = \Phi_S \Phi_R^T$, we can kernelize $f_C(\bm{\alpha}_k,\bm{\beta}_k)$ as follows:

\begin{align*}
	f_C(\bm{\alpha}_k,\bm{\beta}_k)		
&= \bm{\alpha}_k^T \Phi_X^T C \Phi_X \bm{\alpha}_k +
\bm{\beta}_k^T \Phi_Z^T C \Phi_Z \bm{\beta}_k \nonumber+ \bm{\alpha}_k^T \Phi_X^T C \Phi_Z \bm{\beta}_k +
\bm{\beta}_k^T \Phi_Z^T C \Phi_X \bm{\alpha}_k \\
&= \bm{\alpha}_k^T \Phi_X^T \Phi_S \Phi_R^T \Phi_X \bm{\alpha}_k +
\bm{\beta}_k^T \Phi_Z^T \Phi_S \Phi_R^T \Phi_Z \bm{\beta}_k \nonumber \\
&\qquad + \bm{\alpha}_k^T \Phi_X^T \Phi_S \Phi_R^T \Phi_Z \bm{\beta}_k +
\bm{\beta}_k^T \Phi_Z^T \Phi_S \Phi_R^T \Phi_X \bm{\alpha}_k \\
	&= \bm{\alpha}_k^T H_{XX} H_{XZ}^T  \bm{\alpha}_k +
	\bm{\beta}_k^TH_{ZX}H_{ZZ}^T \bm{\beta}_k   \\
	&\qquad  \bm{\alpha}_k^T H_{XX}H_{ZZ}^T   \bm{\beta}_k 
	 + \bm{\beta}_k^T H_{ZX}H_{XZ}^T \bm{\alpha}_k \\		 
	&= [\bm{\alpha}_k^T  \bm{\beta}_k^T ]
	\left[\begin{array}{@{}cc@{}}
	H_{XX}H_{XZ}^T & H_{XX}H_{ZZ}^T\\
	H_{ZX}H_{XZ}^T & H_{ZX}H_{ZZ}^T
  \end{array}\right]
  \left[\begin{array}{@{}c@{}}
  \bm{\alpha}_k \\
  \bm{\beta}_k
   \end{array}\right]\\
   	&= [\bm{\alpha}_k^T  \bm{\beta}_k^T ]
\left[\begin{array}{@{}cc@{}}
H_{XX}\\
H_{ZX}
  \end{array}\right]
    \left[\begin{array}{@{}cc@{}}
H_{XZ}^T & H_{ZZ}^T
   \end{array}\right] 
  \left[\begin{array}{@{}c@{}}
  \bm{\alpha}_k \\
  \bm{\beta}_k
   \end{array}\right]\\
   &=\bm{\theta}_k^T\widetilde{C}\bm{\theta}_k
\end{align*}
For kernelizing $f_D(\bm{\alpha}_k,\bm{\beta}_k)$, it can observed using Eq. (\ref{eqn:f_A}),  the relations in (\ref{eqn:H1}) and  the definition $D = \Phi_R \Phi_S^T$, that $f_D(\bm{\alpha}_k,\bm{\beta}_k) = f_C^T(\bm{\alpha}_k,\bm{\beta}_k)$, as shown below:
\begin{align*}
	f_D(\bm{\alpha}_k,\bm{\beta}_k)		
&= \bm{\alpha}_k^T \Phi_X^T D \Phi_X \bm{\alpha}_k +
\bm{\beta}_k^T \Phi_Z^T D \Phi_Z \bm{\beta}_k \nonumber+ \bm{\alpha}_k^T \Phi_X^T D \Phi_Z \bm{\beta}_k +
\bm{\beta}_k^T \Phi_Z^T D \Phi_X \bm{\alpha}_k \\
&= \bm{\alpha}_k^T \Phi_X^T \Phi_R \Phi_S^T \Phi_X \bm{\alpha}_k +
\bm{\beta}_k^T \Phi_Z^T \Phi_R \Phi_S^T \Phi_Z \bm{\beta}_k \\
&\quad \quad + \bm{\alpha}_k^T \Phi_X^T \Phi_R \Phi_S^T \Phi_Z \bm{\beta}_k +
\bm{\beta}_k^T \Phi_Z^T \Phi_R \Phi_S^T \Phi_X \bm{\alpha}_k \\
&= (\bm{\alpha}_k^T \Phi_X^T \Phi_S \Phi_R^T \Phi_X \bm{\alpha}_k)^T +
(\bm{\beta}_k^T \Phi_Z^T \Phi_S \Phi_R^T \Phi_Z \bm{\beta}_k)^T \\
&\quad \quad + (\bm{\beta}_k^T \Phi_Z^T \Phi_S \Phi_R^T \Phi_X \bm{\alpha}_k)^T + (\bm{\alpha}_k^T \Phi_X^T \Phi_S \Phi_R^T \Phi_Z \bm{\beta}_k)^T
 \\
 &=\Big(\bm{\alpha}_k^T \Phi_X^T C \Phi_X \bm{\alpha}_k+
\bm{\beta}_k^T \Phi_Z^T C \Phi_Z \bm{\beta}_k \nonumber+ \bm{\alpha}_k^T \Phi_X^T C \Phi_Z \bm{\beta}_k +
\bm{\beta}_k^T \Phi_Z^T C \Phi_X \bm{\alpha}_k\Big)^T \\
&=	f^T_C(\bm{\alpha}_k,\bm{\beta}_k)	
\end{align*}
Therefore, it follows that $f_D(\bm{\alpha}_k,\bm{\beta}_k) = \bm{\theta}^T_k\widetilde{C}^T \bm{\theta}_k^T$. {\hfill$\square$}\\

\noindent Based on (\ref{eqn:Num}) and the Lemmas 1,2,3 above,  we finally obtain the following theorem.\\

\noindent\textit{\textbf{Theorem 1:} The denominator term $\mathbf{w}_k^T\Sigma_S\mathbf{w}_k$ in (\ref{eqn:XQDAcost}) can be kernelized as $\mathbf{w}_k^T\Sigma_S\mathbf{w}_k = \bm{\theta}_k^T   \Lambda_S \bm{\theta}_k$, where
\begin{align}
\Lambda_S = (1/n_S)(\widetilde{A} + 			\widetilde{B} - \widetilde{C} - \widetilde{C}^T).
\label{eqn:LambdaS}
\end{align}}\\
This completes the kernelization of the denominator term of (\ref{eqn:XQDAcost}). We next show how the numerator term of (\ref{eqn:XQDAcost}) can be kernelized.\\

\noindent \textbf{Kernelization of numerator} $\mathbf{w}^T_k \Sigma_{D}\mathbf{w}_k$:  
As seen in (\ref{eqn:SigmaD}), the expression for $\Sigma_{D}$ contains $\mathbf{X}$, $\mathbf{Z}$, $\mathbf{s}$ and $\mathbf{r}$. Hence for kernelization, we obtain their representations in the kernel space $\mathcal{F}$ using the kernel function $\phi(\cdot)$ as follows:
\begin{eqnarray}
\Phi_{X} &=& [\phi(\mathbf{x}_1),\dots,\phi(\mathbf{x}_{n_1}),\ldots,\phi(\mathbf{x}_n)] \label{eqn:PhiX} \\
\Phi_{Z} &=& [\phi(\mathbf{z}_1),  \dots, \phi(\mathbf{z}_{m_1}), \dots, \phi(\mathbf{z}_m)] \label{eqn:PhiZ}\\
	\Phi_{s} &=& \sum_{i=1}^n \phi(\mathbf{x}_i), \qquad 	    	\Phi_r = \sum_{i=1}^m \phi(\mathbf{z}_i)  \label{eqn:Phisr}
	\end{eqnarray}
Similar to (\ref{eqn:ABCD}), the covariance matrix $\Sigma_D$ in $\mathcal{F}$ can be expressed using Eq. (\ref{eqn:SigmaD}) as
\begin{equation}
n_D \Sigma_D = \underbrace{m \Phi_X\Phi_X^T}_{U} + \underbrace{n \Phi_Z\Phi_Z^T}_{V} -\underbrace{\Phi_s\Phi_r^T}_{E} -\underbrace{\Phi_r\Phi_s^T}_{P} - n_S \Sigma_S
\label{eqn:MNJL}
\end{equation} 
Then using Eq. (\ref{eqn:rep_theorem2}) and (\ref{eqn:MNJL}), we have
\begin{align}
&\mathbf{w}_k^Tn_D \Sigma_D \mathbf{w}_k =   f_U(\bm{\alpha}_k,\bm{\beta}_k) + f_V(\bm{\alpha}_k,\bm{\beta}_k)\nonumber\\
&\quad -f_E(\bm{\alpha}_k,\bm{\beta}_k)-f_P(\bm{\alpha}_k,\bm{\beta}_k) -\mathbf{w}_k^Tn_S \Sigma_S \mathbf{w}_k \label{eqn:Din}
\end{align} 
where the functions $f_U$, $f_V$, $f_E$ and $f_P$ are of the form
\begin{align}  
f_{\widetilde{Y}}(\bm{\alpha}_k,\bm{\beta}_k) &= \bm{\alpha}_k^T \Phi_{X}^T \widetilde{Y} \Phi_{X} \bm{\alpha}_k +
\bm{\beta}_k^T \Phi_Z^T \widetilde{Y} \Phi_Z \bm{\beta}_k \nonumber\\
 & \qquad + \bm{\alpha}_k^T \Phi_X^T \widetilde{Y} \Phi_Z \bm{\beta}_k +
\bm{\beta}_k^T \Phi_Z^T \widetilde{Y} \Phi_X \bm{\alpha}_k  \label{eqn:f_M}
\end{align}  
for $\widetilde{Y}=U,V,E,P$, which are already defined in (\ref{eqn:MNJL}). 
We next show that each of the terms in (\ref{eqn:Din}) can be expressed as inner products of $\phi(\cdot)$ and hence can be separately kernelized. We have the following two Lemmas.\\

\noindent\textit{\textbf{Lemma 4:} 
$f_U(\bm{\alpha}_k,\bm{\beta}_k)$ and $f_V(\bm{\alpha}_k,\bm{\beta}_k)$ can be kernalized  as 
$f_U(\bm{\alpha}_k,\bm{\beta}_k) = \bm{\theta}_k^T\widetilde{U}\bm{\theta}_k$ and  $f_V(\bm{\alpha}_k,\bm{\beta}_k) = \bm{\theta}_k^T\widetilde{V}\bm{\theta}_k$, where}
\begin{align}
\widetilde{U} &= 
  m \left[\begin{array}{@{}c@{}}
  K_{XX} \\
  K_{ZX}
   \end{array}\right]
   \left[\begin{array}{@{}cc@{}}
     K_{XX}  & K_{XZ}
   \end{array}\right]    \label{eqn:U}\\
\widetilde{V} &= 
  n \left[\begin{array}{@{}c@{}}
  K_{XZ} \\
  K_{ZZ}
   \end{array}\right]
   \left[\begin{array}{@{}cc@{}}
     K_{ZX}  & K_{ZZ}
   \end{array}\right]
   \label{eqn:V}
\end{align}
\noindent\textit{\textbf{Proof:} } 
Using Eq. (\ref{eqn:f_M}), the definition $U=m\Phi_X\Phi_X^T$  and the relations in (\ref{eqn:Kblockmatppty}), we can kernelize $f_{U}(\bm{\alpha}_k,\bm{\beta}_k)$ as follows:
\begin{align*}  
f_{U}(\bm{\alpha}_k,\bm{\beta}_k) &= \bm{\alpha}_k^T \Phi_{X}^T U \Phi_{X} \bm{\alpha}_k +
\bm{\beta}_k^T \Phi_Z^T U\Phi_Z \bm{\beta}_k + \bm{\alpha}_k^T \Phi_X^T U \Phi_Z \bm{\beta}_k +
\bm{\beta}_k^T \Phi_Z^T U \Phi_X \bm{\alpha}_k  \\
	&= \bm{\alpha}_k^T m\Phi_X^T \Phi_X  \Phi_X^T \Phi_X  \bm{\alpha}_k +
	\bm{\beta}_k^T m\Phi_Z^T \Phi_X 	\Phi_X^T \Phi_Z  \bm{\beta}_k  \\
	&\quad + \bm{\alpha}_k^T m\Phi_X^T \Phi_X \Phi_X^T \Phi_Z  \bm{\beta}_k 
	 + \bm{\beta}_k^T m\Phi_Z^T \Phi_X \Phi_X^T \Phi_X  \bm{\alpha}_k \\	
	 	&= \bm{\alpha}_k^T mK_{XX}  K_{XX} \bm{\alpha}_k +
	\bm{\beta}_k^T mK_{ZX}  K_{XZ}  \bm{\beta}_k \\
	&\quad + \bm{\alpha}_k^T mK_{XX}  K_{XZ} \bm{\beta}_k  
	 + \bm{\beta}_k^T mK_{ZX} K_{XX} \bm{\alpha}_k\\
	&= m[\bm{\alpha}_k^T  \bm{\beta}_k^T ]
	\left[\begin{array}{@{}cc@{}}
K_{XX}K_{XX} & K_{XX}K_{XZ}\\
K_{ZX}K_{XX} & K_{ZX}K_{XZ}
  \end{array}\right]
  \left[\begin{array}{@{}c@{}}
  \bm{\alpha}_k \\
  \bm{\beta}_k
   \end{array}\right]\\
   	&= m[\bm{\alpha}_k^T  \bm{\beta}_k^T ]
\left[\begin{array}{@{}cc@{}}
K_{XX}\\
K_{ZX}
  \end{array}\right]    
    \left[\begin{array}{@{}cc@{}}
K_{XX} & K_{XZ}
   \end{array}\right]
  \left[\begin{array}{@{}c@{}}
  \bm{\alpha}_k \\
  \bm{\beta}_k
   \end{array}\right]\\
   &=\bm{\theta}_k^T\widetilde{U}\bm{\theta}_k   
\end{align*}
\noindent  Similarly, $f_{V}(\bm{\alpha}_k,\bm{\beta}_k)$ can also be kernelized using Eq. (\ref{eqn:f_M}), the definition $V=n\Phi_Z\Phi_Z^T$, and the relations in (\ref{eqn:Kblockmatppty}), as follows:
\begin{align*}  
f_V(\bm{\alpha}_k,\bm{\beta}_k) &= \bm{\alpha}_k^T \Phi_{X}^T V \Phi_{X} \bm{\alpha}_k +
\bm{\beta}_k^T \Phi_Z^T V\Phi_Z \bm{\beta}_k + \bm{\alpha}_k^T \Phi_X^T V \Phi_Z \bm{\beta}_k +
\bm{\beta}_k^T \Phi_Z^T V \Phi_X \bm{\alpha}_k  \\
	&= \bm{\alpha}_k^T n\Phi_X^T \Phi_Z  \Phi_Z^T \Phi_X  \bm{\alpha}_k +
	\bm{\beta}_k^T n\Phi_Z^T \Phi_Z  \Phi_Z^T\Phi_Z  \bm{\beta}_k  \\
	&\quad \quad  + \bm{\alpha}_k^T n\Phi_X^T \Phi_Z  \Phi_Z^T \Phi_Z  \bm{\beta}_k 
	 + \bm{\beta}_k^T n\Phi_Z^T \Phi_Z  \Phi_Z^T \Phi_X  \bm{\alpha}_k \\	
	 	&= \bm{\alpha}_k^T nK_{XZ}  K_{ZX} \bm{\alpha}_k +
	\bm{\beta}_k^T nK_{ZZ}  K_{ZZ}  \bm{\beta}_k \\
	&\quad \quad + \bm{\alpha}_k^T nK_{XZ}  K_{ZZ} \bm{\beta}_k 
	 + \bm{\beta}_k^T nK_{ZZ} K_{ZX} \bm{\alpha}_k\\
	&= n 	\left[\begin{array}{@{}cc@{}} 
	\bm{\alpha}_k^T  &  \bm{\beta}_k^T 
	  \end{array}\right]    
	\left[\begin{array}{@{}cc@{}}
K_{XZ}K_{ZX} & K_{XZ}K_{ZZ}\\
K_{ZZ}K_{ZX} & K_{ZZ}K_{ZZ}
  \end{array}\right]
  \left[\begin{array}{@{}c@{}}
  \bm{\alpha}_k \\
  \bm{\beta}_k
   \end{array}\right]\\
   	&= n 	\left[\begin{array}{@{}cc@{}} 
	\bm{\alpha}_k^T  &  \bm{\beta}_k^T 
	  \end{array}\right]
\left[\begin{array}{@{}cc@{}}
K_{XZ}\\
K_{ZZ}
  \end{array}\right]    
    \left[\begin{array}{@{}cc@{}}
K_{ZX} & K_{ZZ}
   \end{array}\right]
  \left[\begin{array}{@{}c@{}}
  \bm{\alpha}_k \\
  \bm{\beta}_k
   \end{array}\right]\\
   &=\bm{\theta}_k^T\widetilde{V}\bm{\theta}_k   
\end{align*}{\hfill$\square$}\\
\noindent\textit{\textbf{Lemma 5:} $f_E(\bm{\alpha}_k,\bm{\beta}_k)$ and $f_P(\bm{\alpha}_k,\bm{\beta}_k)$ can be kernalized  as $f_E(\bm{\alpha}_k,\bm{\beta}_k) = \bm{\theta}_k^T\widetilde{E}\bm{\theta}_k$, and $f_P(\bm{\alpha}_k,\bm{\beta}_k) = \bm{\theta}_k^T\widetilde{E}^T\bm{\theta}_k$ where}
\begin{eqnarray}
\widetilde{E} = 
  \left[\begin{array}{@{}c@{}}
  K_{XX}\\
  K_{ZX}   
   \end{array}\right]
   \left[\begin{array}{@{}c@{}}
   \bm{1}_{n \times m}
      \end{array}\right]
     \left[\begin{array}{@{}cc@{}}
  K_{ZX} & K_{ZZ}
   \end{array}\right]
      \label{eqn:E}
\end{eqnarray}
and  $\bm{1}_{n \times m}$
is an $(n \times m)$ dimensional matrix of ones.

\noindent\textit{\textbf{Proof:} }  For kernelizing $f_E(\bm{\alpha}_k,\bm{\beta}_k)$, we need to express $E=\Phi_s\Phi_r^T$  in terms of $\Phi_X$ and $\Phi_Z$. For that end, we rewrite $\Phi_s$ and $\Phi_r$  based on ( \ref{eqn:Phisr}) as 
	\begin{align}
	\Phi_s &= \sum_{i=1}^n \phi(\mathbf{x}_i) = 
		[\phi(\mathbf{x}_1), \phi(\mathbf{x}_2),\ldots, \phi(\mathbf{x}_n)]
	    \mathbf{1}_n=\Phi_X \mathbf{1}_n\\
	    	\Phi_r &= \sum_{i=1}^m \phi(\mathbf{z}_i) = 
		[\phi(\mathbf{z}_1), \phi(\mathbf{z}_2),\ldots, \phi(\mathbf{z}_m)]
	    \mathbf{1}_m=\Phi_Z \mathbf{1}_m
	\end{align}
where $\mathbf{1}_n$ and $\mathbf{1}_m$ are column vectors of ones having length $n$ and $m$ , respectively. Now based on the definition of $E$, it can be expressed as 
\begin{eqnarray}
	E = \Phi_s\Phi_r^T 
	=\Phi_X \mathbf{1}_{n} \mathbf{1}_{m}^T \Phi_Z^T 
	=\Phi_X \mathbf{1}_{n \times m} \Phi_Z^T \label{eqn:Esolve}
\end{eqnarray}
where $\mathbf{1}_{n \times m}$, is an ${(n \times m)}$ dimensional matrix of ones. Then using Eq. (\ref{eqn:f_M}), (\ref{eqn:Esolve}) and the relations in (\ref{eqn:Kblockmatppty}), we can kernelize $f_E(\bm{\alpha}_k,\bm{\beta}_k)$ as follows:
\begin{align*}  
f_E(\bm{\alpha}_k,\bm{\beta}_k)
 &= \bm{\alpha}_k^T \Phi_{X}^T E \Phi_{X} \bm{\alpha}_k +
\bm{\beta}_k^T \Phi_Z^T E \Phi_Z \bm{\beta}_k + \bm{\alpha}_k^T \Phi_X^T E \Phi_Z \bm{\beta}_k +
\bm{\beta}_k^T \Phi_Z^T E \Phi_X \bm{\alpha}_k  \\
&= \bm{\alpha}_k^T \Phi_X^T (\Phi_X \mathbf{1}_{n \times m} \Phi_Z^T)  \Phi_X  \bm{\alpha}_k +
	\bm{\beta}_k^T \Phi_Z^T (\Phi_X \mathbf{1}_{n \times m} \Phi_Z^T) \Phi_Z  \bm{\beta}_k  \\
	&\qquad \qquad  + \bm{\alpha}_k^T \Phi_X^T(\Phi_X \mathbf{1}_{n \times m} \Phi_Z^T) \Phi_Z  \bm{\beta}_k 
	 + \bm{\beta}_k^T \Phi_Z^T (\Phi_X \mathbf{1}_{n \times m} \Phi_Z^T) \Phi_X  \bm{\alpha}_k \\		
&= \bm{\alpha}_k^T K_{XX} \mathbf{1}_{n \times m} K_{ZX} \bm{\alpha}_k +
	\bm{\beta}_k^T K_{ZX} \mathbf{1}_{n \times m} K_{ZZ}  \bm{\beta}_k \\
	 &\qquad \qquad + \bm{\alpha}_k^T K_{XX} \mathbf{1}_{n \times m} K_{ZZ} \bm{\beta}_k  + \bm{\beta}_k^T K_{ZX} \mathbf{1}_{n \times m} K_{ZX} \bm{\alpha}_k\\
	&= 	\left[\begin{array}{@{}cc@{}} 
	\bm{\alpha}_k^T  &  \bm{\beta}_k^T 
	  \end{array}\right]    
	\left[\begin{array}{@{}cc@{}}
K_{XX}\mathbf{1}_{n \times m}K_{ZX} & K_{XX}\mathbf{1}_{n \times m}K_{ZZ}\\
K_{ZX}\mathbf{1}_{n \times m}K_{ZX} & K_{ZX}\mathbf{1}_{n \times m}K_{ZZ}
  \end{array}\right]
  \left[\begin{array}{@{}c@{}}
  \bm{\alpha}_k \\
  \bm{\beta}_k
   \end{array}\right]\\
   	&=  	\left[\begin{array}{@{}cc@{}} 
	\bm{\alpha}_k^T  &  \bm{\beta}_k^T 
	  \end{array}\right]
\left[\begin{array}{@{}cc@{}}
K_{XX}\\
K_{ZX}
  \end{array}\right]    
  [\mathbf{1}_{n \times m}]
    \left[\begin{array}{@{}cc@{}}
K_{ZX} & K_{ZZ}
   \end{array}\right]
  \left[\begin{array}{@{}c@{}}
  \bm{\alpha}_k \\
  \bm{\beta}_k
   \end{array}\right]\\
   &=\bm{\theta}_k^T\widetilde{E}\bm{\theta}_k   
\end{align*}\\\\
\noindent  For kernelizing $f_P(\bm{\alpha}_k,\bm{\beta}_k)$, it can be seen that 
\begin{eqnarray}
	P = \Phi_r\Phi_s^T 
	=\Phi_Z \mathbf{1}_{m} \mathbf{1}_{n}^T \Phi_X^T 
	=\Phi_Z \mathbf{1}_{m \times n} \Phi_X^T.
\end{eqnarray}
Then, $f_P(\bm{\alpha}_k,\bm{\beta}_k)$ can be kernelized by observing that  $f_P(\bm{\alpha}_k,\bm{\beta}_k) = f_E^T(\bm{\alpha}_k,\bm{\beta}_k)$, as shown below:
\begin{align*}  
f_P(\bm{\alpha}_k,\bm{\beta}_k)
 &= \bm{\alpha}_k^T \Phi_{X}^T P \Phi_{X} \bm{\alpha}_k +
\bm{\beta}_k^T \Phi_Z^T P \Phi_Z \bm{\beta}_k + \bm{\alpha}_k^T \Phi_X^T P \Phi_Z \bm{\beta}_k +
\bm{\beta}_k^T \Phi_Z^T P \Phi_X \bm{\alpha}_k  \\
&= \bm{\alpha}_k^T \Phi_X^T (\Phi_Z \mathbf{1}_{m \times n} \Phi_X^T)  \Phi_X  \bm{\alpha}_k +
	\bm{\beta}_k^T \Phi_Z^T (\Phi_Z \mathbf{1}_{m \times n} \Phi_X^T)  \Phi_Z  \bm{\beta}_k  \\
	&\qquad   \qquad + \bm{\alpha}_k^T \Phi_X^T(\Phi_Z \mathbf{1}_{m \times n} \Phi_X^T)  \Phi_Z  \bm{\beta}_k 
	 + \bm{\beta}_k^T \Phi_Z^T (\Phi_Z \mathbf{1}_{m \times n} \Phi_X^T)   \Phi_X  \bm{\alpha}_k \\			
	 &= (\bm{\alpha}_k^T \Phi_X^T (\Phi_X \mathbf{1}_{n \times m} \Phi_Z^T)  \Phi_X  \bm{\alpha}_k)^T +
	(\bm{\beta}_k^T \Phi_Z^T (\Phi_X \mathbf{1}_{n \times m} \Phi_Z^T) \Phi_Z  \bm{\beta}_k)^T  \\
	&\qquad  \qquad + (\bm{\beta}_k^T \Phi_Z^T (\Phi_X \mathbf{1}_{n \times m} \Phi_Z^T) \Phi_X  \bm{\alpha}_k)^T  +  (\bm{\alpha}_k^T \Phi_X^T(\Phi_X \mathbf{1}_{n \times m} \Phi_Z^T) \Phi_Z  \bm{\beta}_k)^T \\ 
&= \Big[\bm{\alpha}_k^T \Phi_X^T (\Phi_X \mathbf{1}_{n \times m} \Phi_Z^T)  \Phi_X  \bm{\alpha}_k +
	\bm{\beta}_k^T \Phi_Z^T (\Phi_X \mathbf{1}_{n \times m} \Phi_Z^T) \Phi_Z  \bm{\beta}_k   \\
	&\qquad   \qquad + \bm{\alpha}_k^T \Phi_X^T(\Phi_X \mathbf{1}_{n \times m} \Phi_Z^T) \Phi_Z  \bm{\beta}_k  + 	\bm{\beta}_k^T \Phi_Z^T (\Phi_X \mathbf{1}_{n \times m} \Phi_Z^T) \Phi_X  \bm{\alpha}_k \Big]^T\\
 &= f_E^T(\bm{\alpha}_k,\bm{\beta}_k)
\end{align*}\\\\
Then it follows that $f_P(\bm{\alpha}_k,\bm{\beta}_k) = \bm{\theta}_k^T\widetilde{E}^T\bm{\theta}_k $. {\hfill$\square$}\\


\noindent Using Eq. (\ref{eqn:Din}), and the above Lemmas 4 and 5, we get the following theorem.\\\\
\noindent\textit{\textbf{Theorem 2:}
The kernalized form of the denominator term in (\ref{eqn:XQDAcost}) is obtained as $\mathbf{w}_k^T\Sigma_D \mathbf{w}_k =  \bm{\theta}_k^T  \Lambda_D \bm{\theta}_k$
where }
\begin{equation}
\Lambda_D =(1/n_D) ( \widetilde{U} + 			\widetilde{V} - \widetilde{E} - \widetilde{E}^T  - n_S \Lambda_S).
\label{eqn:LambdaD}
\end{equation}\\
Based on Theorem 1 and 2, the kernalized version of the cost function $J(\mathbf{w}_k)$ in (\ref{eqn:XQDAcost}) can now be finally written as
\begin{eqnarray}
J(\bm{\theta}_k) =  \frac{\bm{\theta}_k^T \Lambda_{D}\bm{\theta}_k}{\bm{\theta}_k^T \Lambda_S\bm{\theta}_k}
\label{eqn:KXQDAcost}
\end{eqnarray}
The kernelized cost function $J(\bm{\theta}_k)$ is also of the form of Generalized Rayleigh Quotient. Hence the optimal solutions $\bm{\theta}_k$ that maximize (\ref{eqn:KXQDAcost}) are composed of the eigenvectors corresponding to the $b$ largest eigenvalues of $\Lambda_S^{-1}\Lambda_D$. Similar to XQDA, the dimensionality $b$ of the kXQDA subspace is determined by the number of eigenvectors whose eigenvalues are larger than 1, as it ensures that variance of the dissimilar class $\Sigma_D$ is always higher than the variance of similar class $\Sigma_S$, facilitating effective discrimination between the classes based on difference in variances. \\\\


%

\noindent \textbf{3.3.2   $\quad$  Kernelization of distance metric }\\\\
Next, we kernelize the distance metric $d(\mathbf{x}_i,\mathbf{z}_j)$ in (\ref{eqn:XQDAmetric}). In the kernel space $\mathcal{F}$, the distance metric will be of form
\begin{align}
			d (\Phi(\mathbf{x}_{i}),\Phi(\mathbf{z}_{j})) &= (\Phi(\mathbf{x}_{i})-\Phi(\mathbf{z}_{j}))^T W_{\phi}(\Sigma^{\prime-1}_{S}  - \Sigma^{\prime-1 }_{D})_{+} 
			 W_{\phi}^{T}(\Phi(\mathbf{x}_{i})-\Phi(\mathbf{z}_{j}))\, ,
\label{distmetric_Ker1}
\end{align} 	  
where $\Sigma^{\prime}_S = W_{\phi}^T \Sigma_S W_{\phi}^T$ and $\Sigma^{\prime}_D = W_{\phi}^T \Sigma_D W_{\phi}^T$.\\

\noindent\textit{\textbf{Lemma 6:} The matrices $\Sigma^{\prime}_{S}$ and $\Sigma^{\prime}_{D}$ can be kernalized as $
\Sigma^{\prime}_{S} = \Theta^T \Lambda_S \Theta$, $\; 
\Sigma^{\prime}_{D} = \Theta^T \Lambda_D \Theta,
$
where $\Theta=\left[ \bm{\theta}_1, \bm{\theta}_2, \ldots, \bm{\theta}_b \right]$.}\\

\noindent\textit{\textbf{Proof:} } 
Based on Theorems 1 and 2, it can be seen that, for any general $p, q \in \mathbb{N}$,  the kernelized version of $\mathbf{w}^T_p \Sigma_{D}\mathbf{w}_q$ and $\mathbf{w}^T_p \Sigma_{S}\mathbf{w}_q$ can be written as
\begin{align}
\mathbf{w}^T_p \Sigma_{S}\mathbf{w}_q& = \bm{\theta}_p^T \Lambda_{S}\bm{\theta}_q \label{eqn:wpqS}\\
\mathbf{w}^T_p \Sigma_{D}\mathbf{w}_q &= \bm{\theta}_p^T \Lambda_{D}\bm{\theta}_q \label{eqn:wpqD}
\end{align}
Using the definition of $\Sigma^{\prime}_{S}$, and Eq. (\ref{eqn:wpqD}), we can kernelize  $\Sigma^{\prime}_{S}$ as follows:
\begin{align}
\Sigma^{\prime}_{S} 
&= W_{\phi}^T \Sigma_{S}W_{\phi}\\
&=\left[\begin{array}{@{}cc@{}}
 \mathbf{w}_1^T \\
 \mathbf{w}_2^T\\
 \vdots\\
 \mathbf{w}_b^T
  \end{array}\right]  \Sigma_{S}  \left[ \mathbf{w}_1, \mathbf{w}_2, \ldots, \mathbf{w}_b \right]
  =\left[\begin{array}{@{}cc@{}}
 \bm{\theta}_1^T \\
 \bm{\theta}_2^T\\
 \vdots\\
 \bm{\theta}_b^T
  \end{array}\right]  \Lambda_S  \left[ \bm{\theta}_1, \bm{\theta}_2, \ldots, \bm{\theta}_b \right]\\ 
&=  \Theta^T \Lambda_S \Theta
\end{align}
Similarly, we can can kernelize  $\Sigma^{\prime}_{D}$ using its definition and  Eq. (\ref{eqn:wpqS}) as following: 
\begin{align}
\Sigma^{\prime}_{D} 
&= W^T_{\phi} \Sigma_{D}W_{\phi}\\
&=\left[\begin{array}{@{}cc@{}}
 \mathbf{w}_1^T \\
 \mathbf{w}_2^T\\
 \vdots\\
 \mathbf{w}_b^T
  \end{array}\right]  \Sigma_{D}  \left[ \mathbf{w}_1, \mathbf{w}_2, \ldots, \mathbf{w}_b \right]
  =\left[\begin{array}{@{}cc@{}}
 \bm{\theta}_1^T \\
 \bm{\theta}_2^T\\
 \vdots\\
 \bm{\theta}_b^T
  \end{array}\right]  \Lambda_D  \left[ \bm{\theta}_1, \bm{\theta}_2, \ldots, \bm{\theta}_b \right]\\ 
&=  \Theta^T \Lambda_D \Theta
\end{align}{\hfill$\square$}\\\\

\noindent Using (\ref{eqn:rep_theorem2}), the matrix $W_{\phi}$ can be expressed as
\begin{align}
W_{\phi} = \left[ \mathbf{w}_1, \mathbf{w}_2, ..., \mathbf{w}_b \right]=\bm{\Phi}\left[ \bm{\theta}_1, \bm{\theta}_2, \ldots, \bm{\theta}_b \right] = \bm{\Phi}\Theta
\label{eqn:Wkernel}
\end{align}
Then, using (\ref{eqn:Wkernel}), the initial part of the expression in (\ref{distmetric_Ker1}) can be kernalized as:
\begin{equation}
(\Phi(\mathbf{x}_{i})-\Phi(\mathbf{z}_{j}))^T W_{\phi}     	 
		 =	(K_i - K_j)^T\Theta
		 \label{distmetric_Ker3}
\end{equation} 
where $K_i$ is the $i$th column of the kernel matrix $\mathbf{K}$ in (\ref{eqn:mainK}). \\\\

\noindent Using Lemma 6 and (\ref{distmetric_Ker3}), we finally obtain the following theorem:\\\\
\noindent\textit{\textbf{Theorem 3:}
The kernelized distance metric of kXQDA can be expressed as 
\begin{equation}
d(\Phi(\mathbf{x}_{i}),\Phi(\mathbf{z}_{j})) = 
			(K_i-K_j)^T\Theta \Gamma_{+} \Theta^T (K_i-K_j) 
\label{eqn:KXQDAfinalMahDistM} \\		
\end{equation}
 where $\Gamma = \big[(\Theta^T \Lambda_S \Theta)^{-1} 
 - \Theta^T \Lambda_D \Theta)^{-1}\big]$.}\\

It can be seen that we obtain clean and simplified expressions for k-XQDA as shown in (\ref{eqn:KXQDAcost}) and (\ref{eqn:KXQDAfinalMahDistM}). They have  similar structure compared to the expressions (\ref{eqn:XQDAcost}) and (\ref{eqn:XQDAmetric}) of XQDA. Though our derivations for kernelizing XQDA using (\ref{eqn:SigmaS}) and (\ref{eqn:SigmaD}) is little involved, it should be noted that in our kernelized formulation, there is no requirement of explicit computation of the $nm$ similar/dissimilar class pairs and their outer products for estimating the covariance matrices, which would have been other wise required if (\ref{KISSME_CovMat_calc}) was used for kernelization. Thus our approach achieves a computational reduction of two orders of magnitude. The matrices $\widetilde{A}$, $\widetilde{B}$, $\widetilde{C}$, $\widetilde{U}$, $\widetilde{V}$, and $\widetilde{E}$ required for calculating matrices $\Lambda_D$  and $\Lambda_{S}$  are simplified for fast and efficient computation. They can be easily computed once the matrices $K_{XX}, K_{XZ}, K_{ZZ}, H_{XX}, Z_{ZZ}, H_{XZ}$ and $H_{ZX}$ are obtained.  For the calculation of the eigen system of $\Lambda_S^{-1}\Lambda_D$, we add a small regularizer of $\lambda=10^{-7}$ to the diagonal elements of $\Lambda_S$ to make its estimation more smooth and robust.

 Note that in small sample size case (where $n+m \ll d$) , $\Lambda_S \in \mathbb{R}^{(n+m)\times(n+m)}$ has a much lesser dimension compared to $\Sigma_S \in \mathbb{R}^{d\times d}$ of XQDA. Hence $\Lambda_S$ has lesser number of zero eigen values compared to $\Sigma_S$, making the former better regularizable for inversion. Thus k-XQDA can  handle small sample size (SSS) problem  more efficiently compared to XQDA. Also,  as all other inherent matrices of k-XQDA depends on the number of samples, while that of XQDA depends on the feature dimension, k-XQDA is much faster compared to XQDA. The complete algorithm for k-XQDA is summarized in Algorithm \ref{algo:kxqda}.

\renewcommand{\algorithmicrequire}{\textbf{Input:}}
\renewcommand{\algorithmicensure}{\textbf{Output:}}
\renewcommand{\algorithmiccomment}[1]{// #1}
\begin{algorithm}[t!]
\caption{k-XQDA algorithm}
\label{algo:kxqda}
\begin{algorithmic}[1]
\REQUIRE Training data $\mathbf{X} = (\mathbf{x}_1, \mathbf{x}_2, \ldots,\mathbf{x}_n)$, $\mathbf{Z} = (\mathbf{z}_1, \mathbf{z}_2, \ldots,\mathbf{z}_m)$ with labels $y \in \{1,\ldots,c\}$ 
\ENSURE Distance measure between two samples $\mathbf{x}_i$ and $\mathbf{z}_j$.
\STATE Compute kernel matrices $\mathbf{K}, K_{XX},\;K_{ZZ},\;K_{XZ}, \;K_{ZX}$ using     (\ref{eqn:mainK}), (\ref{eqn:Kblockmatppty}) and (\ref{eqn:Symmppty}).
\STATE Construct matrices $\widetilde{F}$ and $\widetilde{G}$ using (\ref{eqn:Ftilde}) and (\ref{eqn:Gtilde}), based on $y$.
\STATE Compute $\widetilde{A}$ and $\widetilde{B}$ using  (\ref{eqn:Atilde}) and    (\ref{eqn:Btilde}).
\STATE Compute matrices $ H_{XX},\;H_{ZZ},\;H_{XZ}, \;H_{ZX}$ and $\widetilde{C}$ using (\ref{eqn:H2}) and (\ref{eqn:C}).
\STATE Compute $\Lambda_S$ using (\ref{eqn:LambdaS}).
\STATE Compute matrices $\widetilde{U}$, $\widetilde{V}$ and $\widetilde{E}$ using  (\ref{eqn:U}), (\ref{eqn:V}) and (\ref{eqn:E}) respectively.
\STATE Compute $\Lambda_D$ using (\ref{eqn:LambdaD}).
\STATE $\Lambda_S \leftarrow \Lambda_S + \lambda I$ 
\STATE $\Theta=\left[ \bm{\theta}_1, \bm{\theta}_2, \ldots, \bm{\theta}_b \right] \leftarrow $ Eigen vectors of $\Lambda_S^{-1}\Lambda_D$ with eigenvalues  greater than 1.
\STATE Compute $K_i$ and $K_j$ corresponding to $\mathbf{x}_i$ and $\mathbf{z}_j$.
\STATE $\Gamma \leftarrow \big[(\Theta^T \Lambda_S \Theta)^{-1} 
 - \Theta^T \Lambda_D \Theta)^{-1}\big]$. \\	
\STATE Calculate the distance $d(\Phi(\mathbf{x}_{i}),\Phi(\mathbf{z}_{j}))$ using  (\ref{eqn:KXQDAfinalMahDistM}).
\end{algorithmic}
\end{algorithm}

\section{Experiments}
\noindent\textbf{Evaluation Protocol}: 
In re-ID experiments, test set identities are considered unseen during training. Hence following the standard protocol \cite{NK3ML,song:scalableManifold,GOG,SCSP,LOMO,metric_ensembles,colornames}, the dataset identities are divided equally into half forming the training set and the other half forming the test set. For training, each person is considered as one distinct class. For testing, the test images from one view form the query set and the rest forms the gallery set. The queries are matched against the gallery and a ranked list is obtained based on the matching score. Rank-N accuracy is calculated as the probability of true match occurring in the first N search results. The above procedure is repeated 10 times and the average performance is evaluated. \\\\
\noindent\textbf{Datasets}: We use four standard datasets including CUHK01\cite{CUHK01}, PRID450S\cite{PRID450S}, GRID\cite{GRID1} and PRID2011\cite{PRID2011}, which have small size training set for our experiments. They contain 971, 450, 250 and 200 persons, respectively, captured from two non-overlapping camera views. Each person has one image in each view, except the CUHK01 dataset, which has two images in each view. For CUHK01, we use both single-shot as well as multi-shot settings. The gallery of GRID and PRID2011 datasets have additional 775 and 549 images, respectively, which are of different identities from the query set and act as distractors.  \\ \\
\noindent\textbf{Features and Parameters}: For each person image, we use standard feature descriptors including WHOS\cite{LisantiPAMI14}, LOMO\cite{LOMO} and GOG\cite{GOG}. The LOMO and GOG are of dimensions 26,960 and 27,622 respectively. The WHOS feature is of two type, one with 2960 and the other with 5138 dimensions. We refer the first as WHOS* and the second as $\text{WHOS}^{\dagger}$. We also use a new feature descriptor named $\text{LOMO}^{\dagger}$, which is the LOMO feature obtained without using Retinex \cite{LOMO} transformation, to make use of of color diversity. 
Re-ID datasets have large variation in illumination and background. Hence for  k-XQDA, we use specific features and kernel functions for each dataset, to better model their inherent characteristics. We use RBF or polynomial kernel for k-XQDA.\\ 

\noindent\textbf{Method of Comparison}: We conduct our experiments using only the given training data. There are some re-ID methods that use external supervision (like pre-trained networks on other datasets or  auxiliary data like human pose, attributes or body part segmentation obtained using external trained systems) and post-processing (re-ranking) of the trained models using the test data. 
No such external supervision or post-processing is considered in our study and hence a direct comparison of our results with such methods is not advisable. However, we list them in separate rows for completeness.

\begingroup
\setlength{\tabcolsep}{10pt}
\begin{table}[t]
\small
\begin{center}
\resizebox{0.75\columnwidth}{!}{%
\begin{tabular}[t]{l|cccc}
\hline
Methods & r=1 & r=5 & r=10 & r=20\\
\hline\hline
\small WHOS* + XQDA  & 33.16&	53.01&	61.57&	70.43\\
WHOS* + k-XQDA  & \color{red}\textbf{43.75}	&\color{red}\textbf{67.30}	&\color{red}\textbf{76.16}	&	\color{red}\textbf{84.12}\\
\hline
\small $\text{WHOS}^{\dagger}$ + XQDA  & 37.61	&58.06&	66.62	&75.14\\
$\text{WHOS}^{\dagger}$ + k-XQDA  & \color{red}\textbf{52.45}	&\color{red}\textbf{76.43}&	\color{red}\textbf{84.60}	&\color{red}\textbf{90.93}\\
\hline
LOMO + XQDA  & 50.00&	75.32&	83.40	&	89.51\\
LOMO + k-XQDA  & \color{red}\textbf{54.43}	&\color{red}\textbf{79.63}&	\color{red}\textbf{86.45}	&	\color{red}\textbf{92.25}\\
\hline
GOG + XQDA  & 57.89	&79.15&	86.25	&	92.14\\
GOG + k-XQDA  & \color{red}\textbf{62.23}	&\color{red}\textbf{83.09}&	\color{red}\textbf{89.46}	&	\color{red}\textbf{94.43}\\
\hline
\end{tabular}
}
\end{center}
\caption{Comparison of k-XQDA with the baseline method XQDA on CUHK01 dataset, using single-shot settings.}
\label{table:BaselineXQDA}
\end{table}
\endgroup

\begingroup
\setlength{\tabcolsep}{10pt}
\begin{table}[t]
\small
\begin{center}
\resizebox{0.75\columnwidth}{!}{%
\begin{tabular}[t]{l c ccccc}
\hline
Methods & Ref &r=1 & r=5 & r=10 & r=20\\
\hline\hline
\small 
ITML &\cite{ITML}& 22.60& 40.60 &50.40 &61.50\\
LMNN &\cite{LMNN1}& 42.30& 61.50 &70.50 &79.20 \\
LFDA &\cite{LFDA:CVPR}& 44.67&	67.37&	76.05&	83.31\\
kLFDA &\cite{rPcca}& 46.67&	72.38&	81.96	&	89.01\\
MFA & \cite{rPcca}&42.55	&69.76	&80.45&		88.42\\
KISSME & \cite{KISSME} &41.87	&68.95&	79.21	&87.58\\
MLAPG & \cite{MLAPG}& 51.79 & 75.29& 82.54 &  89.41\\
NFST &\cite{Zheng:nfst} &40.04	&62.21	&71.67&	80.08\\
KNFST & \cite{Zheng:nfst} &52.80	&77.28	&84.97	&	91.07\\
XQDA&	\cite{LOMO} &50.00	&75.32	&83.40	&	89.51\\
\hline
k-XQDA  & Ours & \color{red}\textbf{54.43}&	\color{red}\textbf{79.63}	&\color{red}\textbf{86.45}	&	\color{red}\textbf{92.25}\\
\hline
\end{tabular}
}
\end{center}
\caption{Comparison of k-XQDA with baseline metric learning methods using the same LOMO feature  on CUHK01 dataset (single-shot).}
\label{table:Baselinek-XQDA}
\end{table}
\endgroup

\subsection{Comparison with Baselines}
As k-XQDA is the kernalized version of XQDA, we first compare its performance against XQDA. We extensively evaluate using multiple feature descriptors including WHOS*, WHOS, LOMO and GOG, and the results are shown in Table \ref{table:BaselineXQDA}. k-XQDA consistently outperforms XQDA with high margin, at all ranks. For WHOS* descriptor, k-XQDA attains an improvement of 10.59\% at rank-1 and 14.29\% at rank-5, against XQDA. Similarly for $\text{WHOS}^{\dagger}$ descriptor, k-XQDA outperforms XQDA by 14.84\% at rank-1 and 18.37\% at rank-5. For LOMO and GOG feature descriptors, a rank-1 performance boost of 4.43\% and 4.34\% are respectively obtained by k-XQDA. Thus, independent of the feature descriptor used, k-XQDA has superior performance than XQDA. The results signify that, with the benefit of kernels, k-XQDA is able to learn efficient non-linear features than XQDA for handling the  high non-linearity in person appearances across cameras. 

Next we compare the performance of k-XQDA against other state-of-the-art metric learning methods including MLAPG\cite{MLAPG}, NFST\cite{Zheng:nfst}, KNFST\cite{Zheng:nfst}, KISSME\cite{KISSME}, LFDA\cite{LFDA:CVPR} and kLFDA\cite{rPcca}.  We conduct experiments using the same LOMO feature descriptor on CUHK01 dataset, and the results are shown in Table \ref{table:Baselinek-XQDA}. It can be seen that k-XQDA outperforms all the compared metric learning methods. Note that KNFST\cite{Zheng:nfst} and kLFDA\cite{rPcca} are kernel based methods and our kernel based method k-XQDA attains the highest performance. The experiment also confirms the inferences drawn in \cite{Zheng:nfst} and \cite{rPcca} that kernel based methods are very crucial for handling non-linearity in person re-identification.

\begingroup
\setlength{\tabcolsep}{10pt}
\subsection{Comparison with State-of-the-art}
\begin{table}[t]
\begin{center}
\small
\resizebox{0.65\columnwidth}{!}{%
\begin{tabular}{lccc}
\hline
Methods & Rank1 &  Rank10 & Rank20 \\
\hline \hline
KISSME\cite{KISSME} & 15.00  & 39.00 & 52.00 \\
LMNN\cite{LMNN1} & 10.00  & 30.00 & 42.00 \\
ITML\cite{ITML} & 12.00  & 36.00 & 47.00 \\
Mahal\cite{PRID450S} & 16.00 & 41.00 &51.00\\
RPLM \cite{RPLM} & 15.00  &42.00 &54.00\\
TPC \cite{TCP} & 22.00	& 	47.00	&	57.00\\
XQDA\cite{LOMO}&27.80	&	59.60	&	71.20\\
KNFST \cite{Zheng:nfst} & 29.80		& 66.00	&	76.50 \\
\textit{l}1-graph \cite{UlGraph} & 30.10  &- & -\\
SBAL\cite{SBAL} & 32.40 &-&-\\
\textbf{k-XQDA}  & \color{red}\textbf{35.30}	&	\color{red}\textbf{72.10}	&\color{red}\textbf{81.70}\\
\hline
*MuDeep\cite{MuDeep} & 17.90  & 45.90 & 55.40\\
*MetricEnsembles\cite{metric_ensembles} & 17.90  & 50.00 & 62.00\\
*PTGAN \cite{PTGAN}& 33.50 &  71.50& - \\
*MC-PPMN\cite{MCPPMN}&  34.00 & 69.00& -\\
\hline
\end{tabular}
}
\end{center}
\caption{Performance comparison with state-of-the-art results on PRID2011 dataset.  The methods with a * signifies post processing / external supervision based methods.}
\label{table:PRID2011}
\end{table}
\endgroup

\noindent\textbf{Experiments with PRID2011 dataset:}
PRID2011 is a challenging dataset with very small training data. We use GOG features for this dataset. 
As seen in Table \ref{table:PRID2011}, our proposed methods k-XQDA attains competitive performance against the state-of-the-art results for all ranks.
We clearly outperform all the deep learning based methods including MuDeep\cite{MuDeep}. The deep learning methods PTGAN\cite{PTGAN} and MC-PPMN\cite{MCPPMN} uses auxiliary supervision
 while our method have better performance, even without using any extra information, except the given training images.\\

\noindent\textbf{Experiments with CUHK01 dataset:}
Concatenated LOMO, $\text{LOMO}^{\dagger}$ and GOG are used as the features. 
 For \textit{single-shot} settings, where every person has only one image in each view, the results are shown in Table \ref{table:CUHKM1}. kXQDA attains the best results at all ranks. 
 Note that we even outperformed the body pose based auxiliary supervised deep learning method PN-GAN\cite{PNGAN}.  For \textit{multi-shot} experiments also, we attain competitive performance against state-of-the-art methods, as shown in Table \ref{table:CUHKM2}. This additionally signifies that our methods can also handle multiple images per class, efficiently. \\

\begingroup
\setlength{\tabcolsep}{10pt}
\begin{table}[t]
\small
\begin{center}
\begin{tabular}{lccc}
\hline
Methods &  Rank1 & Rank10 & Rank20 \\
\hline\hline
\small 
MLFL\cite{midlevel}  & 34.30& 65.00&75.00\\
XQDA\cite{LOMO}  & 50.00		&83.40		&89.51\\
KNFST\cite{Zheng:nfst}  &52.80 	&84.97		&91.07\\

TPC \cite{TCP}  &53.70		&91.00&96.30\\
CAMEL\cite{CAMEL}  & 57.30 &-	&-\\
GOG\cite{GOG}	 & 57.89 &  86.25 &  92.14\\
WARCA\cite{WARCA} & 58.34 &- &-\\
MVLDML+\cite{MVLDML} & 61.37 & 88.88 & 93.85\\
\textbf{k-XQDA} &  \color{red}\textbf{67.77}	&	\color{red}\textbf{92.23}	&	\color{red}\textbf{95.94}\\
\hline
*Semantic\cite{Symantic}		&	32.70& 64.40 &76.30	\\
*MetricEnsemble\cite{metric_ensembles}  & 53.40 &  84.40&   90.50\\
*Quadruplet\cite{Beyond:triplet_loss}  & 62.55& 89.71		&-\\
*PN-GAN\cite{PNGAN} & 67.65 & 91.82 & -\\
\hline
\end{tabular}
\end{center}
\caption{Comparison with state-of-the-art results on CUHK01 dataset using  single-shot  settings. The methods with a * signifies post processing / external supervision based methods.}
\label{table:CUHKM1}
\end{table}
\endgroup

\begingroup
\setlength{\tabcolsep}{10pt}
\begin{table}[h!]
\small
\begin{center}
\begin{tabular}{lccc}
\hline
Methods  & Rank1 & Rank10 & Rank20 \\
\hline\hline
\textit{l}1-Graph\cite{UlGraph}  & 50.10&-&-\\
GCT\cite{GCT} & 61.90 &  87.60 &92.80 \\
XQDA\cite{LOMO} & 61.98 	&89.30		&93.62\\
CAMEL\cite{CAMEL}  &62.70 &-&-\\
MLAPG\cite{MLAPG}  & 64.24 &90.84&94.92\\
SSSVM\cite{SSSVM} &65.97&-&-\\
KNFST\cite{Zheng:nfst}&66.07&		91.56&		95.64\\
GOG\cite{GOG}  & 67.28		&91.77		&95.93\\
IRS(LOMO)\cite{IRS} & 68.39 & 92.60 & 96.20 \\
\textbf{k-XQDA} &  \color{red}\textbf{76.30}	&	\color{red}\textbf{95.39}	&	\color{red}\textbf{98.15}\\
\hline
*DGD\cite{DGD} & 66.60& -&-\\
*OLMANS\cite{OnlineNegSamples}  & 68.44&		92.67&	95.88\\
*SHaPE\cite{SHaPE} & 76.00 & -&-\\
\hline
\end{tabular}
\end{center}
\caption{Comparison with state-of-the-art results on CUHK01 dataset using multi-shot settings.}
\label{table:CUHKM2}
\end{table}
\endgroup

\begingroup
\setlength{\tabcolsep}{10pt}
\begin{table}[ht]
\small
\begin{center}
\begin{tabular}[c]{lccc}
\hline
Methods &  Rank1 & Rank10 & Rank20 \\
\hline\hline
WARCA\cite{WARCA} & 24.58 & - & -\\
SCNCD\cite{colornames}		&	41.60	&	79.40	&	87.80	\\
CSL\cite{CSL}		&	44.40	&	82.20	&	89.80	\\
TMA\cite{TMA}		&	52.89	&	85.78	&	93.33	\\
k-KISSME\cite{kKISSME} & 53.90  & 88.80 & 94.50\\
GCT\cite{GCT} & 58.40&  84.30&89.80\\
KNFST\cite{Zheng:nfst} &	 59.47	&	91.96	&	96.53\\
XQDA\cite{LOMO}		&	59.78	&	90.09	&	95.29	\\
SSSVM\cite{SSSVM}		&	60.49	&	88.58	&	93.60	\\
MC-PPMN\cite{MCPPMN} & 62.22 &  93.56 & -\\
MVLDML+\cite{MVLDML} & 66.80 & 94.80 & 97.7\\
GOG+XQDA\cite{GOG}		&	68.00 	&	94.36	&	97.64	\\
\textbf{k-XQDA} &\color{red}\textbf{73.16}	&	\color{red}\textbf{95.91}	& \color{red}\textbf{98.44} \\
\hline
*Semantic\cite{Symantic}		&	44.90	&	77.50	&	86.70	\\
*SSM\cite{song:scalableManifold}	&	72.98	&	96.76	&	99.11	\\
\hline
\end{tabular}
\end{center}
\caption{Comparison with state-of-the-art results on  PRID450S dataset. }
\label{tab:PRID450Sall}
\end{table}
\endgroup

\noindent\textbf{Experiments with PRID450S dataset:}
We use concatenated GOG+LOMO+$\text{LOMO}^{\dagger}$ as the features in our methods. As shown in Table \ref{tab:PRID450Sall}, we attain competitive performance with state-of-the-art results. We also outperform the post-processing based method SSM\cite{song:scalableManifold}. It is a re-ranking method that utilize gallery data, while our method uses only the training data. Hence it can be expected that any general re-ranking method like SSM can be used on top of our method to further increase our performance.\\

\begingroup
\setlength{\tabcolsep}{10pt}
\begin{table}[h!]
\small
\begin{center}
\begin{tabular}[c]{lccc}
\hline
Methods &  Rank1 & Rank10 & Rank20 \\
\hline\hline
\small MtMCML\cite{MtMCML}		&	14.08	&	45.84	&	59.84	\\
KNFST\cite{Zheng:nfst}   & 14.88	&	41.28	&	50.88 \\
PolyMap\cite{ExPolyFeatMap}		&	16.30	&	46.00	&	57.60	\\
XQDA\cite{LOMO}		&	16.56	&	41.84	&	52.40	\\
MLAPG\cite{MLAPG}		&	16.64	&	41.20	&	52.96	\\
KEPLER\cite{KEPLER}		&	18.40	&	50.24	&	61.44	\\
DR-KISS\cite{DR-KISS}		&	20.60	&	51.40	&	62.60	\\
SSSVM\cite{SSSVM}  & 22.40	& 51.28	&	61.20\\
SCSP\cite{SCSP}		&	24.24	&	54.08	&	65.20	\\
GOG\cite{GOG}	&	24.80	&	58.40	&	68.88	\\
\textbf{k-XQDA} & \color{red}\textbf{27.28}	&	\color{red}\textbf{58.96}	&	\color{red}\textbf{69.12}\\
\hline
*SSDAL\cite{SSDAL}		&	22.40	&	48.00	&	58.40	\\
*SSM\cite{song:scalableManifold}  & 27.20	&	61.12	&	70.56\\
*OL-MANS\cite{OnlineNegSamples}  & 30.16 &	49.20 &	59.36\\
\hline
\end{tabular}
\end{center}
\caption{Comparison with state-of-the-art results on  GRID dataset.}
\end{table}
\endgroup

\noindent\textbf{Experiments with GRID dataset}
GRID is a very challenging dataset. We use concatenated GOG, LOMO and $\text{LOMO}^{\dagger}$ as the features. 
Our method has competitive performance against the state-of-the-art methods. Though OLMANS\cite{OnlineNegSamples} have slightly higher performance at rank-1, we outperform it in rank-10 and 20. Moreover, OLMANS needs to compute a separate secondary metric for every query image, making it more computationally intensive, while our method is computationally efficient.\\

\section{Conclusion}
In this paper we proposed a new kernel based non-linear cross-view similarity metric learning approach that can learn non-linear transformations and handle complex non-linear appearance change of persons across camera views. Using kernel based mapping to a higher dimensional space, a discriminative subspace as well as a Mahalanobis metric is learned by discriminating the similar class and dissimilar class based on their ratio of variances. Through our rigorous derivations, we obtain simplified expressions for the distance metric, making it computationally very efficient and fast. The method handles small size training data for practical person re-identification systems and better solves the small sample size problem. Extensive experiments on four benchmark datasets shows that the proposed method achieves competitive performance against many state-of-the-art methods.\\

\noindent \textbf{Acknowledgment.} This research work is supported under Visvesvaraya PhD Scheme by Ministry of Electronics and Information Technology (MeitY), Government of India. 

\bibliographystyle{splncs04}
\bibliography{egbibOrigICCV2019}

\begin{thebibliography}{10}
\providecommand{\url}[1]{\texttt{#1}}
\providecommand{\urlprefix}{URL }
\providecommand{\doi}[1]{https://doi.org/#1}

\bibitem{ImprDeep}
Ahmed, E., Jones, M., Marks, T.K.: An improved deep learning architecture for
  person re-identification. CVPR  (2015)

\bibitem{NK3ML}
Ali, T.M.F., Chaudhuri, S.: Maximum margin metric learning over discriminative
  nullspace for person re-identification. ECCV  (2018)

\bibitem{SemiNK3ML}
Ali, T.M.F., Chaudhuri, S.: A semi-supervised maximum margin metric learning
  approach for small scale person re-identification. ICCVw  (2019)

\bibitem{MFML}
Ali, T.M.F., Patel, K.K., Velmurugan, R., Chaudhuri, S.: Multiple kernel fisher
  discriminant metric learning for person re-identification. ICVGIP  (2018)

\bibitem{song:scalableManifold}
Bai, S., Bai, X., Tian, Q.: Scalable person re-identification on supervised
  smoothed manifold. CVPR  (2017)

\bibitem{SHaPE}
Barman, A., Shah, S.K.: Shape: A novel graph theoretic algorthm for making
  consensus-based decisions in person re-identification systems. ICCV  (2017)

\bibitem{SCSP}
Chen, D., Yuan, Z., Chen, B., Zheng, N.: Similarity learning with spatial
  constraints for person re-identification. CVPR  (2016)

\bibitem{ExPolyFeatMap}
Chen, D., Yuan, Z., Hua, G., Zheng, N., Wang, J.: Similarity learning on an
  explicit polynomial kernel feature map for person re-identification. CVPR
  (2015)

\bibitem{Beyond:triplet_loss}
Chen, W., Chen, X., Zhang, J., Huang, K.: Beyond triplet loss: a deep
  quadruplet network for person re-identification. CVPR  (2017)

\bibitem{TCP}
Cheng, D., Gong, Y., Zhou, S., Wang, J., Zheng, N.: Person re-identification by
  multi-channel parts-based cnn with improved triplet loss function. CVPR
  (2016)

\bibitem{ITML}
Davis, J.V., Kulis, B., Jain, P., Sra, S., Dhillon, I.S.: Information-theoretic
  metric learning. ICML  (2007)

\bibitem{LDML}
Guillaumin, M., Verbeek, J., Schmid, C.: Is that you? metric learning
  approaches for face identification. ICCV  (2009)

\bibitem{RPLM}
Hirzer, M., Roth, P.M., Kostinger, M., Bischof, H.: Relaxed pairwise learned
  metric for person re-identification. ECCV  (2012)

\bibitem{PRID2011}
Hirzer, M., Beleznai, C., Roth, P.M., Bischof, H.: Person re-identification by
  descriptive and discriminative classification. Image analysis

\bibitem{WARCA}
Jose, C., Fleuret, F.: Scalable metric learning via weighted approximate rank
  component analysis. ECCV  (2016)

\bibitem{UlGraph}
Kodirov, E., Xiang, T., Fu, Z., Gong, S.: Person re-identification by
  unsupervised \textit{l}1 graph learning. ECCV  (2016)

\bibitem{KISSME}
Köstinger, M., andP. Wohlhart, M.H., Roth, P.M., Bischof, H.: Large scale
  metric learning from equivalence constraints. CVPR  (2012)

\bibitem{CUHK01}
Li, W., Zhao, R., Wang, X.: Human reidentification with transferred metric
  learning. ACCV  (2012)

\bibitem{LOMO}
Liao, S., Hu, Y., Zhu, X., Li, S.Z.: Person re-identification by local maximal
  occurrence representation and metric learning. CVPR  (2015)

\bibitem{MLAPG}
Liao, S., Li, S.Z.: Efficient psd constrained asymmetric metric learning for
  person re-identification. ICCV  (2015)

\bibitem{LisantiPAMI14}
Lisanti, G., Masi, I., Bimbo, A.D.: Person re-identification by iterative
  re-weighted sparse ranking. IEEE TPAMI  (2014)

\bibitem{SBAL}
Liu, W., Chang, X., Chen, L., Yang, Y.: Semi-supervised bayesian attribute
  learning for person re-identification. AAAI  (2018)

\bibitem{GRID1}
Loy, C.C., Xiang, T., Gong, S.: Multi-camera activity correlation analysis.
  CVPR  (2009)

\bibitem{MtMCML}
Ma, L., Yang, X., Tao, D.: Person re-identification over camera networks using
  multi-task distance metric learning. IEEE TIP  (2014)

\bibitem{MCPPMN}
Mao, C., Li, Y., Zhang, Y., Zhang, Z., Li, X.: Multi-channel pyramid person
  matching network for person re-identification. AAAI  (2018)

\bibitem{TMA}
Martinel, N., Das, A., Micheloni, C., Chowdhury, A.K.R.: Temporal model
  adaptation for person reidentification. ECCV  (2016)

\bibitem{KEPLER}
Martinel, N., Micheloni, C., Foresti, G.L.: Kernelized saliency-based person
  re-identification through multiple metric learning. IEEE TIP  (2015)

\bibitem{GOG}
Matsukawa, T., Okabe, T., Suzuki, E., Sato, Y.: Hierarchical gaussian
  descriptor for person re-identification. CVPR  (2016)

\bibitem{kKISSME}
Nguyen, B., De~Baets, B.: Kernel distance metric learning using pairwise
  constraints for person re-identification. IEEE TIP  (2018)

\bibitem{metric_ensembles}
Paisitkriangkrai, S., Shen, C., van~den Hengel, A.: Learning to rank in person
  re-identification with metric ensembles. CVPR  (2015)

\bibitem{LFDA:CVPR}
Pedagadi, S., Orwell, J., Velastin, S., Boghossian, B.: Local fisher
  discriminant analysis for pedestrian re-identification. CVPR  (2013)

\bibitem{MuDeep}
Qian, X., Fu, Y., Jiang, Y.G., Xiang, T., Xue, X.: Multi-scale deep learning
  architectures for person re-identification. In: ICCV (2017)

\bibitem{PNGAN}
Qian, X., Fu, Y., Xiang, T., Wang, W., Qiu, J., Wu, Y., Jiang, Y.G., Xue, X.:
  Pose-normalized image generation for person re-identification. ECCV  (2018)

\bibitem{PRID450S}
Roth, P.M., Hirzer, M., Koestinger, M., Beleznai, C., Bischof, H.: Mahalanobis
  distance learning for person re-identification. In Person Re-Identification
  (2014)

\bibitem{CSL}
Shen, Y., Lin, W., Yan, J., Xu, M., Wu, J., Wang, J.: Person re-identification
  with correspondence structure learning. ICCV  (2015)

\bibitem{Symantic}
Shi, Z., Hospedales, T.M., Xiang, T.: Transferring a semantic representation
  for person re-identification and search. CVPR  (2015)

\bibitem{SSDAL}
Su, C., Zhang, S., Xing, J., Gao, W., Tian, Q.: Deep attributes driven
  multi-camera person re-identification. ECCV  (2016)

\bibitem{DR-KISS}
Tao, D., Guo, Y., Song, M., Li, Y., Yu, Z., Tang, Y.Y.: Person
  re-identification by dual-regularized kiss metric learning. IEEE TIP  (2016)

\bibitem{SCNN}
Varior, R.R., Haloi, M., Wang., G.: Gated siamese convolutional neural network
  architecture for human reidentification. ECCV  (2016)

\bibitem{SLSTM}
Varior, R.R., Shuai, B., Lu, J., Xu, D., Wang, G.: A siamese long short-term
  memory architecture for human reidentification. ECCV  (2016)

\bibitem{IRS}
Wang, H., Zhu, X., Gong, S., Xiang, T.: Person re-identification in identity
  regression space. IJCV  (2018)

\bibitem{PTGAN}
Wei, L., Zhang, S., Gao, W., Tian, Q.: Person transfer gan to bridge domain gap
  for person re-identification. CVPR  (2018)

\bibitem{LMNN1}
Weinberger, K.Q., Blitzer, J., Saul, L.K.: Distance metric learning for large
  margin nearest neighbor classification. NIPS  (2006)

\bibitem{DGD}
Xiao, T., Li, H., Ouyang, W., Wang, X.: Learning deep feature representations
  with domain guided dropout for person re-identification. CVPR  (2016)

\bibitem{rPcca}
Xiong, F., Gou, M., Camps, O., Sznaier, M.: Person re-identification using
  kernel-based metric learning methods. ECCV  (2014)

\bibitem{MVLDML}
Yang, X., Wang, M., Tao, D.: Person re-identification with metric learning
  using privileged information. IEEE TIP  (2018)

\bibitem{colornames}
Yang, Y., Yang, J., Yan, J., Liao, S., Yi, D., Li, S.Z.: Salient color names
  for person re-identification. ECCV  (2014)

\bibitem{CAMEL}
Yu, H.X., Wu, A., Zheng, W.S.: Cross-view asymmetric metric learning for
  unsupervised person re-identification. ICCV  (2017)

\bibitem{Zheng:nfst}
Zhang, L., Xiang, T., Gong, S.: Learning a discriminative null space for person
  re-identification. CVPR  (2016)

\bibitem{SSSVM}
Zhang, Y., Li, B., 1, H.L., 2, A.I., Ruan, X.: Sample-specific svm learning for
  person re-identification. CVPR  (2016)

\bibitem{SpindleNet}
Zhao, H., Tian, M., Sun, S., Shao, J., Yan, J., Yi, S., Wang, X., Tang, X.:
  Spindle net: Person re-identification with human body region guided feature
  decomposition and fusion. CVPR  (2017)

\bibitem{midlevel}
Zhao, R., Ouyang, W., Wang., X.: Learning mid-level filters for person
  re-identification. CVPR  (2014)

\bibitem{Reranking:kreciprocal}
Zhong, Z., Zheng, L., Cao, D., Li, S.: Re-ranking person re-identification with
  k-reciprocal encoding. CVPR  (2017)

\bibitem{OnlineNegSamples}
Zhou, J., Yu, P., Tang, W., Wu, Y.: Efficient online local metric adaptation
  via negative samples for person re-identification. ICCV  (2017)

\bibitem{GCT}
Zhou, Q., Fan, H., Zheng, S., Su, H., Li, X., Wu, S., Ling, H.: Graph
  correspondence transfer for person re-identification. AAAI  (2018)

\end{thebibliography}

\end{document}